\documentclass[letterpaper]{article} 
\usepackage{aaai2026}  
\usepackage{times}  
\usepackage{helvet}  
\usepackage{courier}  
\usepackage[hyphens]{url}  
\usepackage{graphicx} 
\usepackage{natbib}  
\usepackage{caption} 
\usepackage{algorithm}
\usepackage{algorithmic}

\usepackage{amsmath}
\usepackage{amsfonts}

\usepackage{multirow}
\usepackage{booktabs}

\usepackage{amsthm}

\newtheorem{theorem}{Theorem}
\newtheorem{lemma}{Lemma}

\usepackage{cuted}
\usepackage{amssymb}

\usepackage{newfloat}
\usepackage{listings}
\DeclareCaptionStyle{ruled}{labelfont=normalfont,labelsep=colon,strut=off} 
\floatstyle{ruled}
\newfloat{listing}{tb}{lst}{}
\floatname{listing}{Listing}
\title{Enhancing Rotation-Invariant 3D Learning with Global Pose Awareness and Attention Mechanisms}
\author {
    Jiaxun Guo\textsuperscript{\rm 1},
    Manar Amayri\textsuperscript{\rm 1},
    Nizar Bouguila\textsuperscript{\rm 1},
    Xin Liu\textsuperscript{\rm 2},
    Wentao Fan\textsuperscript{\rm 3} \footnote{Corresponding author.}
}
\affiliations {
    \textsuperscript{\rm 1}CIISE, Concordia University, Montreal, Canada\\
    \textsuperscript{\rm 2}Department of Computer Science and Technology, Huaqiao University, Xiamen, China\\
    \textsuperscript{\rm 3}Guangdong Provincial/Zhuhai Key Laboratory IRADS and Department of Computer Science, Beijing Normal-Hong Kong Baptist University, Zhuhai, China\\
    jiaxun.guo@mail.concordia.ca, \{manar.amayri, nizar.bouguila\}@concordia.ca, xliu@hqu.edu.cn, wentaofan@bnbu.edu.cn \\ 
}

\usepackage{bibentry}

\begin{document}
 
\maketitle

\begin{abstract}
Recent advances in rotation-invariant (RI) learning for 3D point clouds typically replace raw coordinates with handcrafted RI features to ensure robustness under arbitrary rotations. However, these approaches often suffer from the loss of global pose information, making them incapable of distinguishing geometrically similar but spatially distinct structures. We identify that this limitation stems from the restricted receptive field in existing RI methods, leading to \textit{Wing–tip feature collapse}, a failure to differentiate symmetric components (e.g., left and right airplane wings) due to indistinguishable local geometries. To overcome this challenge, we introduce the Shadow-informed Pose Feature (SiPF), which augments local RI descriptors with a globally consistent reference point (referred to as the “shadow”) derived from a learned shared rotation. This mechanism enables the model to preserve global pose awareness while maintaining rotation invariance. We further propose Rotation-invariant Attention Convolution (RIAttnConv), an attention-based operator that integrates SiPFs into the feature aggregation process, thereby enhancing the model’s capacity to distinguish structurally similar components. Additionally, we design a task-adaptive shadow locating module based on the Bingham distribution over unit quaternions, which dynamically learns the optimal global rotation for constructing consistent shadows. Extensive experiments on 3D classification and part segmentation benchmarks demonstrate that our approach substantially outperforms existing RI methods, particularly in tasks requiring fine-grained spatial discrimination under arbitrary rotations.
\end{abstract}


\begin{links}
    \link{Code}{https://github.com/jiaxunguo/EnRI-GAM}
\end{links}

\section{Introduction}

The rapid advancement of 3D sensing technologies has led to the widespread adoption of point cloud data in various 3D computer vision tasks, including shape classification, part segmentation, and object recognition \cite{guo2020deep, afham2022crosspoint, duan2023condaformer, jin2023context}. While recent learning-based models \cite{qi2017pointnet++, wang2019dynamic, liu2019relation} have achieved notable success on pre-aligned datasets, their generalization to real-world scenarios remains limited. A common and often unrealistic assumption in these models is that input objects are consistently aligned in a canonical orientation. However, in practical applications such as autonomous driving, robotics, and augmented reality, 3D objects naturally appear under arbitrary rotations. Models that lack true \textit{rotation-invariant} (RI) capabilities experience substantial performance degradation when confronted with such orientation variations, even when extensively trained with rotation-augmented data. This limitation fundamentally restricts the deployment of point cloud models in unconstrained environments.

\begin{figure}[t]
\centering
    \includegraphics[width=.45\textwidth]{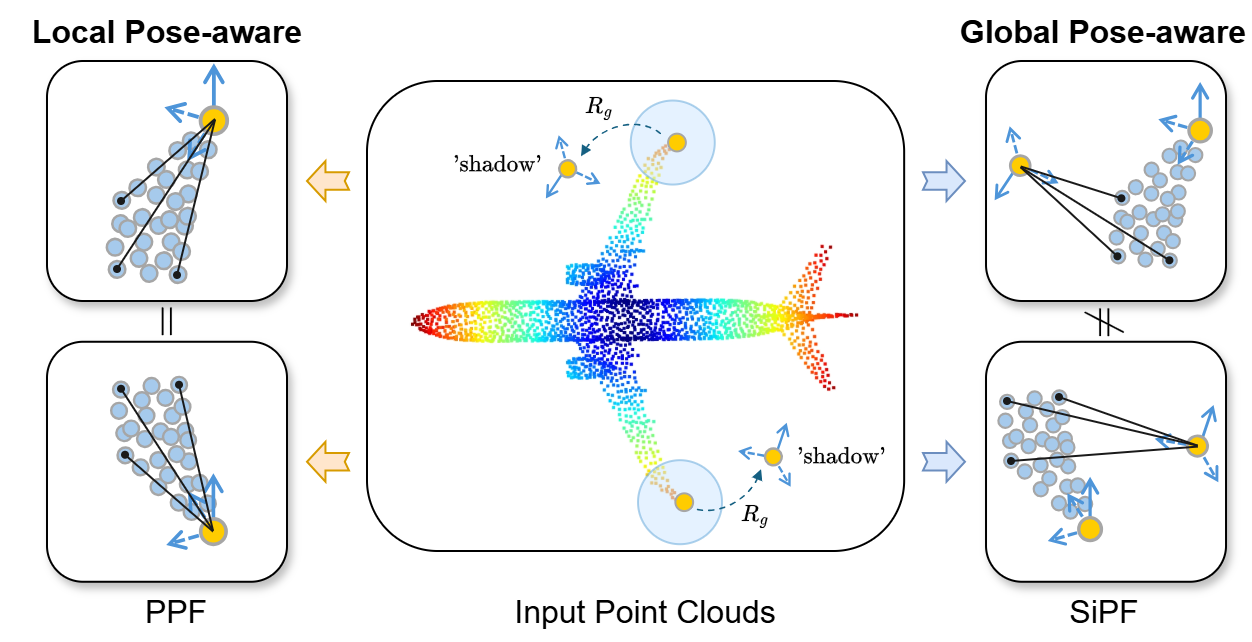}
    \caption{Illustration of pose ambiguity caused by symmetric local structures. Traditional Point Pair Features (PPF) capture only local geometric relationships and thus yield identical representations for symmetric regions with similar local geometry (e.g., left and right wing tips). In contrast, the proposed SiPF incorporates global pose awareness by introducing a `shadow' reference point derived from a shared rotation matrix $R_g$ to incorporate pose-aware context, enabling consistent and discriminative RI representations across arbitrary orientations. }
    \label{fig_idea}
\end{figure}

To overcome this issue, a growing body of research has focused on designing RI models by replacing raw 3D coordinates with handcrafted or learned geometric features that are invariant to global rotations \cite{xu2021sgmnet, gu2022enhanced, li2021rotation}. These approaches demonstrate improved robustness to orientation changes by leveraging local descriptors such as point pair features or RI tensors. However, a critical drawback remains: by discarding absolute coordinate information, these methods also lose global pose context, which is essential for differentiating spatially distinct but geometrically similar regions. For instance, in the case of an airplane model, symmetric components such as the left and right wings may be indistinguishable when only local geometry is considered—regardless of rotation invariance. This leads to what we refer to as the \textit{wing-tip feature collapse} phenomenon. Recent attempts to mitigate this issue involve expanding the receptive field and incorporating pairwise relationships within local neighbourhoods to capture broader context while maintaining RI properties \cite{chen2022devil, zhang2024risurconv}. Despite these advances, existing methods still struggle with structural ambiguity in repetitive or symmetric components, primarily due to their limited capacity to encode spatial distinctiveness without access to a consistent global reference.

We argue that the inherent limitation of the receptive field in current RI methods is a fundamental bottleneck. Specifically, for geometrically similar regions, such as opposing airplane wings, the local features (even aggregated from larger neighbourhoods) lack the discriminative information necessary to encode their distinct global poses. As shown in Figure \ref{fig_idea}, conventional RI descriptors fail to preserve relative global positioning, resulting in ambiguous representations that impede fine-grained recognition tasks like part segmentation. Furthermore, these limitations explain the suboptimal performance of many RI pipelines, which focus on preserving local geometric consistency but neglect global relational cues critical for downstream tasks.

To address these challenges, in this work we propose a novel feature representation termed \textit{Shadow-informed Pose Feature} (SiPF) that augments local RI features with global pose cues while preserving rotation invariance. Inspired by the classical point pair feature (PPF) \cite{drost2010model}, SiPF introduces a globally consistent shadow reference for each point, derived via a learned shared rotation. This ``shadow" serves as an anchor point for encoding relative position in a globally consistent manner. Specifically, we introduce a task-adaptive shadow locating module that estimates a shared optimal rotation via the Bingham distribution over unit quaternions. Using the learned rotation, each point is paired with its shadow to construct SiPF descriptors that retain local geometric relations while incorporating global spatial awareness. We further design a novel attention-based operator, termed \textit{Rotation-invariant Attention Convolution} (RIAttnConv), that integrates SiPF into a cross-attention mechanism fused with convolutional operations. This design enables dynamic aggregation of spatially-aware features, thereby improving the model’s ability to distinguish structurally similar but spatially distinct components. We validate our method on multiple benchmark datasets across classification and part segmentation tasks. Experimental results demonstrate that our framework consistently outperforms state-of-the-art RI methods, particularly in challenging settings that require fine-grained spatial discrimination under arbitrary orientations.

Our main contributions are summarized as follows:
\begin{itemize}
    \item We identify the restricted receptive field as a key limitation in current RI methods and address it by introducing the SiPF, which incorporates global pose context while maintaining rotation invariance.

    \item We develop RIAttnConv, a novel convolutional module that integrates SiPF through an attention-augmented feature aggregation strategy, significantly enhancing the discriminative capacity of RI representations.

    \item We design a task-adaptive shadow locating module that learns a globally consistent reference rotation using the Bingham distribution over unit quaternions, enabling dynamic and robust shadow construction.

\end{itemize}

\section{Related Works}

\subsection{Rotation-Sensitive 3D Point Cloud Analysis}\;
PointNet \cite{qi2017pointnet} is a pioneering work in point cloud analysis, introducing a straightforward yet effective framework for directly processing raw 3D point sets in tasks such as classification and segmentation. To address its limitations in capturing local geometric structures, PointNet++ \cite{qi2017pointnet++} extends this architecture with hierarchical feature learning over local neighbourhoods. Subsequent works, such as PointCNN \cite{li2018pointcnn} and DGCNN \cite{wang2019dynamic}, further enhance local context modeling through specialized convolutional and dynamic graph-based operations, respectively. Despite these advances, a shared limitation among these approaches is their sensitivity to input orientation. Because they operate directly on absolute coordinates, their performance tends to degrade significantly when encountering arbitrarily rotated inputs, which limits their applicability in real-world scenarios.

\subsection{Rotation-Robust Methods}\;
To mitigate the sensitivity to rotations, early approaches utilize principal component analysis (PCA) to align input point clouds to a canonical orientation. For instance, Li et al. \cite{li2021closer} address pose ambiguity through hybrid canonicalization strategies. However, such methods often rely heavily on data augmentation or heuristic approximations, failing to achieve true rotation invariance. 
An alternative direction involves equivariant neural networks that encode predictable transformations of features under rotations via structured linear mappings \cite{deng2021vector, poulenard2021functional, fuchs2020se}. These models preserve pose information while improving rotation robustness. Nonetheless, maintaining equivariance typically imposes strict architectural constraints, such as enforcing linear compositions in encoders \cite{deng2021vector, jing2020learning}, which can limit model expressiveness and hinder learning capacity. 
Recently, constructing RI features within local neighbourhoods has emerged as a prevalent strategy. Most of these methods employ handcrafted features derived from geometric priors such as pairwise distances and inter-point angles \cite{li2021rotation, xu2021sgmnet}. For instance, RI-CNN \cite{zhang2019rotation} and PaRI-Conv \cite{chen2022devil} design 4D and 8D local descriptors, respectively, to enhance geometric encoding. However, these designs often sacrifice global contextual information, thereby constraining their ability to capture complex spatial semantics \cite{zhang2019rotation}. To alleviate this drawback, several methods have attempted to expand the receptive field or increase connectivity. ClusterNet \cite{chen2019clusternet}, for example, constructs a densely connected $k$NN graph (with $k=80$) to retain richer relational information. LocoTrans \cite{chen2024local} enhances local RI features by leveraging representations derived from an equivariant backbone. 

Recent advances such as RIGA~\cite{yu2024riga}, and rotation-invariant transformers~\cite{yu2023rotation} show that incorporating global context or attention is beneficial for rotation-robust point matching. Attention-based sampling strategies~\cite{wu2023attention, wu2025samble} further highlight the importance of globally informed feature construction. Building on these insights, our approach explicitly integrates global pose information with local geometric encoding, yielding more expressive RI representations and consistently strong performance across rotation-robust benchmarks.

\section{Problem Definition and Background}\label{sec:prob_def}

\noindent
\textbf{Rotation-Invariant Functions.}\;
Let a point cloud be denoted by $P=\{p_1,\dots,p_N\}\in\mathbb{R}^{N\times3}$ and let $R\!\in\!SO(3)$ represents an arbitrary 3D rotation matrix. The rotated point cloud is thus $P_{R}=P\,R$. A function $\Phi$ is RI if and only if:
\begin{equation}
\Phi\!\left(P_{R}\right)=\Phi(P), \quad \forall\,R\in SO(3).
\label{eq:ri}
\end{equation}

\noindent
\textbf{Receptive Field in 3D Point Cloud Convolutions.}\;
Given a point cloud $P=[p_1,\dots,p_N]\!\in\!\mathbb{R}^{N\times3}$ with associated input features
$X=[x_1,\dots,x_N]\!\in\!\mathbb{R}^{N\times c_{\text{in}}}$,  
consider a reference point $p_r$ with its local neighbourhood denoted by $\mathcal{N}(p_r)$.  
The output feature at $p_r$, written as part of $X'=[x'_1,\dots,x'_N]\!\in\!\mathbb{R}^{N\times c_{\text{out}}}$, is typically obtained by aggregating transformed features of its neighbours:
\begin{equation}
f(p_r)=
\bigwedge_{j\in\mathcal{N}(p_r)}
W_j^r \cdot h \bigl(p_j \bigr).
\label{eq:conv_basic}
\end{equation}
where $h:\mathbb{R}^{3}\!\rightarrow\!\mathbb{R}^{c}$ is a nonlinear point-encoding function,  
$W_j^r$ denotes the learnable dynamic kernel weight associated with the pair $(p_r, p_j)$, and $\bigwedge$ represents an aggregation operator such as MAX, SUM, or AVG. 

Let the local patch around point $p_j$ be defined as $\Omega(p_j)=\{p_k \mid k\in\mathcal{N}(p_j)\}$.  
Since $h$ may also depend on the neighbourhood of $p_j$, the effective receptive field at $p_r$ becomes the union of its neighbours' patches:
\begin{equation}
\Gamma(p_r)=
\bigcup_{j\in\mathcal{N}(p_r)}
\Omega(p_j).
\label{eq:rf}
\end{equation}

To make the dependence on the receptive field explicit, Equation~\ref{eq:conv_basic} can be reformulated as:
\begin{equation}
f(p_r | \Gamma(p_r))=
\bigwedge_{j\in\mathcal{N}(p_r)}
W_j^r \cdot h \bigl(p_j | \Omega(p_j) \bigr).
\label{eq:conv_ext}
\end{equation}

\noindent\textbf{Limitations of the Receptive Field in Existing RI Methods.}\;
Consider an airplane point cloud and let $p_{\mathrm{left}}, p_{\mathrm{right}}\in P$ be the two wing-tip points. Due to their geometric symmetry, there exists a rotation
\(R_{\mathrm{sym}}\in SO(3)\) such that:  
\begin{equation}
\Omega(p_{\mathrm{right}})=\Omega(p_{\mathrm{left}})\,R_{\mathrm{sym}} .
\label{eq:re_batch}
\end{equation}

\noindent
\textit{Patch–Swapping Transformation.}\;
For any reference point \(p_r\) with neighbourhood \(\mathcal N(p_r)\), we define a transformation on the receptive field as:
\begin{equation}
\label{eq:ps_trans}
T\bigl(\Gamma(p_r)\bigr)=
\bigcup_{j\in\mathcal N(p_r)}
\Omega(p_j)\,R_j ,
\end{equation}
where each \(R_j\in SO(3)\) maps the local patch \(\Omega(p_j)\) to its geometrically equivalent counterpart (e.g.\ \(R_{\mathrm{sym}}\) swaps the left and right wing-tip patches). Applying Equation~\ref{eq:conv_ext} to the transformed data yields:
\begin{equation}
f\!\bigl(p_r\,\bigl|\,T(\Gamma(p_r))\bigr)=
\bigwedge_{j\in\mathcal N(p_r)}
W_j^r\cdot h\!\bigl(p_j\,\bigl|\,\Omega(p_j)R_j\bigr).
\label{eq:conv_ext_rot}
\end{equation}
Since \(h\) is rotation-invariant by definition (Equation~\ref{eq:ri}), we have \(h(p_j\,|\,\Omega(p_j)R_j)=h(p_j\,|\,\Omega(p_j))\). Therefore, Equations~\ref{eq:conv_ext} and \ref{eq:conv_ext_rot} are equivalent, leading to:
\begin{equation}
f\!\bigl(p_r\,\bigl|\,\Gamma(p_r)\bigr)=
f\!\bigl(p_r\,\bigl|\,T(\Gamma(p_r))\bigr).
\label{eq:conv_eq}
\end{equation}

\noindent
\textit{Wing-Tip Feature Collapse.}\;
By setting \(p_r=p_{\mathrm{left}}\) and choosing \(T\) to swap the left and right wing-tip patches, Equation~(\ref{eq:conv_eq}) simplifies to: \(f(p_{\mathrm{left}})=f(p_{\mathrm{right}})\). This result implies that the network assigns identical features to symmetric yet globally opposite regions. As a consequence, crucial pose information (e.g. distinction between left and right) is lost, introducing ambiguity in downstream tasks and limiting the discriminative power of the learned representations.

\section{The Proposed Method}
To address the global pose ambiguity that arises among geometrically equivalent local patches, we propose a novel RI point descriptor, termed SiPF. It augments each point with a global pose-aware reference, known as the \emph{shadow}, obtained by applying a shared rotation to the entire point cloud. This augmentation allows local patches to encode relative pose information without compromising rotation invariance, thereby enabling the network to distinguish globally distinct yet locally similar structures. 

\begin{figure*}[htbt]
\centering
    \includegraphics[width=1\textwidth]{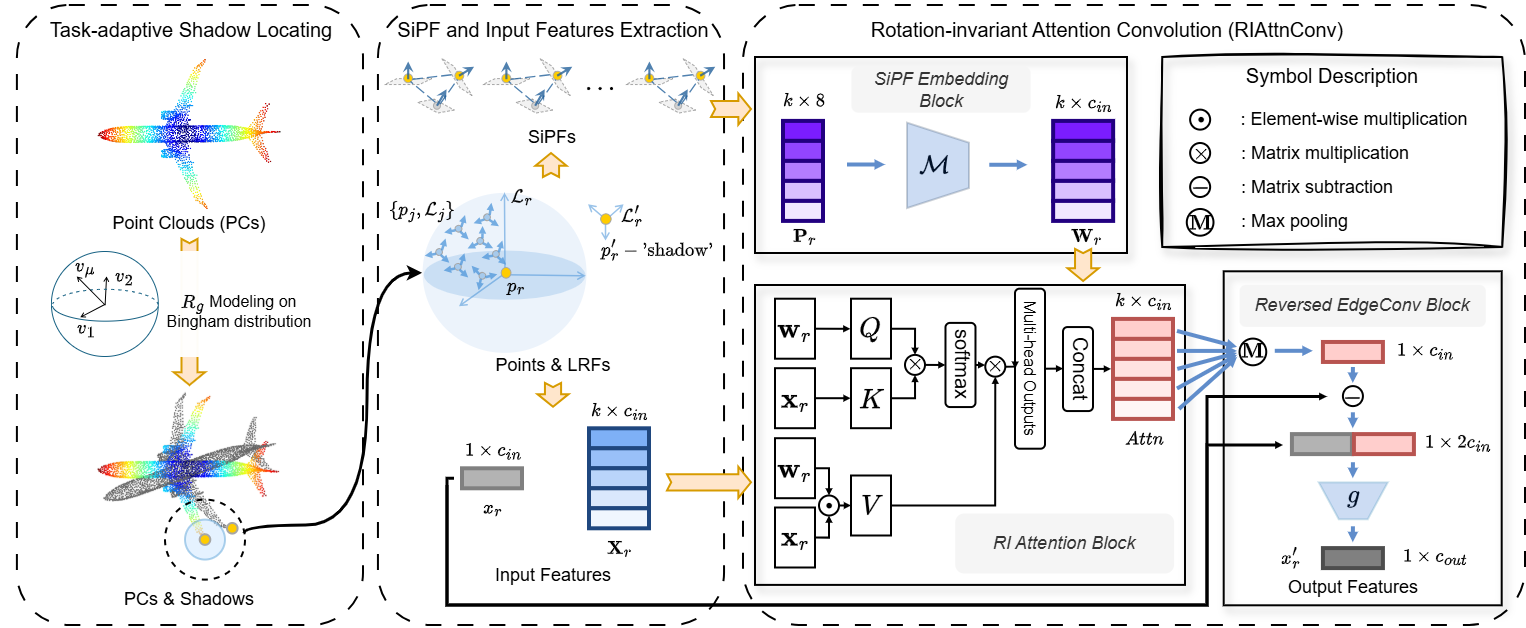}
    \caption{Overview of the proposed SiPF and RIAttnConv pipeline. Our framework consists of three main components: (1) Task-adaptive Shadow Locating estimates a globally consistent reference rotation $R_g$ by modeling unit quaternions with a Bingham distribution, generating `shadow' points for each input; (2) SiPF and Input Feature Extraction constructs LRFs $\{\mathcal{L}_j\}^k_{j=1}$ and computes SiPFs $\mathbf{P}_r$ by combining local geometric descriptors with global shadow-based pose differences; (3) RIAttnConv embeds SiPFs to produce adaptive kernel weights $\mathbf{W}_r$, which are used in a RI attention block to guide feature aggregation. A reversed EdgeConv module then fuses neighbourhood and center features $x_r$ to produce the final output $x_r'$.}
    \label{fig_full}
\end{figure*}

Building upon SiPF, we further introduce RIAttnConv, a rotation-invariant convolutional operator that incorporates shadow-informed relative pose information into the feature aggregation process. Specifically, for a given reference point \(p_r\), the output feature is computed as:
\begin{equation}
x'_r = \bigwedge_{j \in \mathcal{N}(p_r)} \mathcal{A}ttn \left( \{\mathcal{W}(\mathcal{P}_r^j) \cdot x_j \}_{j=1}^{k} \right),
\label{eq: idea_abs}
\end{equation}
where \(\mathcal{P}_r^j\) encodes the relative pose among the reference point \(p_r\), its shadow \(p_r'\), and neighbour \(p_j\). The kernel weight \(W_j^r\) adapts dynamically to the global context introduced by \(p_r'\), allowing the model to distinguish symmetric structures based on their global orientation. As a result, \(W_j^r\) in Equations~(\ref{eq:conv_ext}) and~(\ref{eq:conv_ext_rot}) varies under local patch rotations, enabling the attention module \(Attn(\cdot)\) to differentiate identical local geometries in distinct global poses while preserving overall rotation invariance. The full architectural pipeline of our method is illustrated in Figure \ref{fig_full}.

\subsection{Shadow-Informed Pose Feature}

\noindent\textbf{LRF Construction.}\;
Local Reference Frame (LRF) is a fundamental component for constructing RI representations. While the encoding on the relative pose \(\mathcal{P}_r^j\) between point pairs, the LRF \(\mathcal{L}_r\) serves as a local coordinate system independently constructed at each point \(p_r \in P\). Formally, \(\mathcal{L}_r = [\partial_r^1, \partial_r^2, \partial_r^3] \in \mathbb{R}^{3 \times 3}\) consisting of three orthonormal basis vectors, typically derived from two direction vectors \(e_r^1, e_r^2\) via Gram–Schmidt orthonormalization:
\begin{equation}
\partial^1_r = e^1_r, \quad 
\partial^3_r = \frac{\partial^1_r \times e^2_r}{\|\partial^1_r \times e^2_r\|}, \quad 
\partial^2_r = \partial^3_r \times \partial^1_r.
\end{equation}

Recent works \cite{chen2022devil, zhang2024risurconv} adopted a robust and purely local LRF construction strategy, wherein the vector from the barycenter of point neighbours to the point itself defines one axis. Specifically, the barycenter is computed as \(m_r = \frac{1}{k}\sum_{j=1}^{k} p_j^r\), where \(p_j^r\) are the \(k\)-nearest neighbours of \(p_r\), and the axis \(e_r^2\) is defined as \(\overrightarrow{p_r m_r} = m_r - p_r\). In this work, we adopt this LRF construction strategy based on local geometry (\(e_r^1 = n_r, e_r^2 = \overrightarrow{p_r m_r}\)) as input features \( X\) to ensure essential robustness and consistency in real-world point clouds.

\smallskip
\noindent\textbf{SiPF Extraction.}\;
To describe the relative pose between the LRFs of a reference point \(p_r\) and its neighbour \(p_j\), we design our RI feature based on the standard Point Pair Feature (PPF) \cite{drost2010model}, a 4D vector constructed from the relative position and the primary LRF axes \(\partial_r^1\), \(\partial_j^1\) of the two points:
\begin{equation}
\begin{aligned}
\text{PPF}(p_r, p_j)& = (\|d\|_2,\; \cos(\alpha_1),\; \cos(\alpha_2),\; \cos(\alpha_3)), \\
\alpha_1 = \angle(\partial_r^1, &d), \quad 
\alpha_2 = \angle(\partial_j^1, d), \quad 
\alpha_3 = \angle(\partial_r^1, \partial_j^1),
\end{aligned}
\label{eq:ppf}
\end{equation}
where \(d = p_j - p_r\) and \(\angle(\cdot,\cdot)\) denotes the angle between two vectors. While PPF is sufficient to represent the pairwise pose, it becomes ambiguous in the context of multiple neighbours. As shown in prior works \cite{deng2018ppf, chen2022devil}, for two neighbours \(p_j\) and \(p_j'\) with identical normals and equal distances to \(p_r\), if both lie on a circle around the axis \(\partial_r^1\), their PPF values with respect to \(p_r\) will be identical, i.e., \(\text{PPF}(p_r, p_j) = \text{PPF}(p_r, p_j')\), despite having distinct spatial arrangements. 

To resolve this ambiguity and incorporate global pose awareness, we introduce the \emph{shadow} point \(p_r'\) of \(p_r\), derived from a shared global rotation. We then define the \emph{Shadow-informed Point Pair Feature} (SiPPF) as:
\begin{equation}
\text{SiPPF}(p_r, p_r', p_j) = 
\frac{\text{PPF}(p_r, p_r') - \text{PPF}(p_j, p_r')}{\|\text{PPF}(p_r, p_r') - \text{PPF}(p_j, p_r')\|_2},
\label{eq:sippf}
\end{equation}
where \(\text{PPF}(p_r, p_r')\) and \(\text{PPF}(p_j, p_r')\) encode the relative geometry between each point and the shadow. We normalize the difference using the \(\ell_2\)-norm to emphasize the pose difference of \(p_j\) under the shared global reference.

Finally, we concatenate the local geometric encoding with the global-aware difference to obtain our 8D \textit{Shadow-informed Pose Feature} (SiPF):
\begin{equation}
\mathcal{P}_r^j = \left(\text{PPF}(p_r, p_j),\; \text{SiPPF}(p_r, p_r', p_j)\right) \in \mathbb{R}^8.
\label{eq:sipf}
\end{equation}

\smallskip
\noindent\textbf{Task-Adaptive Shadow Locating.}\;
As previously discussed, a shared global rotation is applied to generate the shadow point \(p_r'\), which serves as a consistent reference for global-relative feature extraction. However, in certain geometric configurations, the shadow may fail to enhance the discriminability of local features due to inherent ambiguities. Two representative failure cases are outlined below:
\begin{itemize}
    \item[(1)] When the shadow \(p_r'\) lies along the primary LRF axis \(\partial_r^1\), and the primary axes \(\partial_r^1\) and \(\partial_{r'}^1\) coincide, the SiPF degenerates to standard PPF due to unresolved degrees of freedom of $p_j$.
    \item[(2)] If the global rotation coincides with the local rotation path \(R_j\) used in Equation \ref{eq:ps_trans}, the resulting SiPF only cancels the internal degrees of freedom of \(p_j\), but fails to distinguish the receptive field \(\Omega(p_j)\) from its rotated variant \(\Omega(p_j)R_j\).
\end{itemize}
These cases highlight the necessity of a task-adaptive, ambiguity-resilient global rotation strategy. To account for the ambiguity in selecting \(\mathbf{R}_g\) over the entire dataset, we model the uncertainty using a \emph{Bingham distribution} on the hypersphere \(S^3\) of unit quaternions \cite{bingham1974antipodally}. Let \(q\) represent candidate quaternions corresponding to potential global rotations. The Bingham probability density function is defined as:
\begin{equation}
\mathcal{B}(q \mid \mathbf{V}, \mathbf{\Lambda}) = \frac{1}{F(\mathbf{\Lambda})} \exp\left(q^\top \mathbf{V} \mathbf{\Lambda} \mathbf{V}^\top q\right),
\label{eq:bingham}
\end{equation}
where \(\mathbf{V} \in \mathbb{R}^{4 \times 4}\) is an orthogonal matrix, and \(\mathbf{\Lambda} = \text{diag}(\lambda_1, \lambda_2, \lambda_3, 0)\) is a diagonal matrix with \(\lambda_1 \leq \lambda_2 \leq \lambda_3 < 0\). The matrix \(\mathbf{V}\) defines the mode and dispersion axes, while \(\mathbf{\Lambda}\) controls the strength of dispersion along these axes. \(F(\mathbf{\Lambda})\) is the normalization constant. To connect \(\mathbf{R}_g\) with the learned mode of the Bingham distribution, we extract the mode vector in \(\mathbf{V}\) as the optimal global rotation candidate for generating \(p_r' = p_r\mathbf{R}_g\) at each iteration.

To ensure that the selected global rotation aligns with the task-specific objectives, we embed the optimization of \(\mathbf{R}_g\) into the end-to-end training. Specifically, we jointly optimize the rotation with network parameters by minimizing a composite loss that integrates both the task objective and the probabilistic alignment with the Bingham distribution:
\begin{equation}
\mathcal{L}_{\text{total}} = \mathcal{L}_{\text{task}} + \delta \cdot \sqrt{\left(\mathcal{L}_{\text{bingham}} - 0.1\cdot\mathcal{L}_{\text{task}}\right)^2},
\end{equation}
where \(\mathcal{L}_{\text{bingham}}\) denotes the negative log-likelihood of the Bingham distribution evaluated at the quaternion \(q = \text{Quat}(\mathbf{R}_g)\), and \(\delta\) is a balancing hyperparameter, empirically set to 0.8. The subtraction term encourages to maintain consistency between the uncertainty captured by the Bingham model and the alignment needs of the downstream task.

\subsection{The RIAttnConv Operator}

To enhance RI feature learning, we propose the RIAttnConv operator. Unlike prior approaches that rely only on relative pose for determining kernel weights, RIAttnConv employs an attention mechanism that adaptively weights neighbour features based on both geometric cues and feature similarity. This design enables the model to capture broader contextual dependencies while facilitating effective aggregation.

\smallskip
\noindent\textbf{RI Attention.}\;
Given a reference feature \(x_r \in \mathbb{R}^{c_{\text{in}}}\) and its \(k\)-nearest neighbouring features \(\{x_j\}_{j=1}^k\), we define the stacked neighbour feature matrix as \(\mathbf{X}_r = [x_1, \dots, x_k]^\top \in \mathbb{R}^{k \times c_{\text{in}}}\). Similarly, the corresponding SiPF descriptors are stacked as \(\mathbf{P}_r = [\mathcal{P}_r^{1}, \dots, \mathcal{P}_r^{k}]^\top \in \mathbb{R}^{k \times 8}\). RIAttnConv constructs a scaled dot-product attention mechanism using both \(\mathbf{X}_r\) and \(\mathbf{P}_r\):
\begin{equation}
\begin{aligned}
\text{Attention}&(Q, K, V) = \text{Softmax}\left(\frac{Q K^\top}{\sqrt{c_{in}}}\right) V, \\
& Q = \mathbf{W}_r, \quad
K = \mathbf{X}_r, \quad
V = \mathbf{W}_r\cdot \mathbf{X}_r,
\end{aligned}
\label{eq:attention}
\end{equation}
where $\mathbf{W}_r = [W_1^{r}, \dots, W_k^{r}]^\top \in \mathbb{R}^{k\times c_{in}}$, and $W_j^r = \mathcal{M}(\mathcal{P}_r^j)$ as $\mathcal{W}(\mathcal{P}_r^j)$ in Equation (\ref{eq: idea_abs}).  $\mathcal{M}$ is a Multi-layer Perceptron (MLP) that maps the relative pose $\mathcal{P}_r^j$ into the respective dimension.

This design enables the \(\text{Softmax}(\cdot)\) in Equation~(\ref{eq:attention}) to produce a \(k \times k\) attention matrix over the \(k\)-nearest neighbours, enhancing contextual sensitivity. Compared to prior methods \cite{chen2022devil, zhou2021adaptive}, this formulation offers a favourable trade-off between computational complexity and expressive power. A detailed efficiency analysis is provided in ablation study (see Table~\ref{tab:as_ric}).

\smallskip
\noindent\textbf{Reversed EdgeConv.}\;
To retain the feature information of the reference point \(x_r\), we propose a reversed variant of EdgeConv \cite{wang2019dynamic}, where neighbour features are first aggregated and then fused with the reference feature. Specifically, given the attention mechanism, we compute:
\begin{equation}
\begin{aligned}
\mathcal{A}&ttn \left(\{\mathcal{W}(\mathcal{P}_r^j) \cdot x_j \}_{j=1}^{k} \right) = \text{Attention}(Q, K, V), \\
&\hat{x}_r = \text{MAX}_{j \in \mathcal{N}(p_r)} \mathcal{A}ttn \left( \{\mathcal{W}(\mathcal{P}_r^j) \cdot x_j \}_{j=1}^{k} \right),
\end{aligned}
\end{equation}
where \(\mathcal{A}ttn\) is the output from the RI attention module defined in Equation~(\ref{eq:attention}). Then, the updated feature \(x_r'\) is obtained by fusing \(\hat{x}_r\) and the original reference feature \(x_r\) as:
\begin{equation}
x_r' = g\left((\hat{x}_r - x_r) \oplus x_r\right),
\end{equation}
where \(g(\cdot)\) denotes a one-layer MLP and \(\oplus\) represents feature concatenation.

\subsection{Network Architecture}

To validate the generalizability and effectiveness of the proposed RIAttnConv operator, we integrate it into two representative backbone architectures widely adopted in point cloud learning: DGCNN \cite{wang2019dynamic} for classification, and AdaptConv \cite{zhou2021adaptive} for segmentation.

\smallskip
\noindent\textbf{Shape Classification.}\;
We adopt the DGCNN backbone and replace all EdgeConv layers with RIAttnConv. The $k$-nearest neighbour graph is constructed in Euclidean space at each feature layer, with \(k=20\) as the neighbourhood size.

\smallskip
\noindent\textbf{Shape Part Segmentation.}\;
For part segmentation, we follow the AdaptConv-based design. Specifically, we stack four RIAttnConv layers followed by a graph convolution layer. Three intermediate pooling stages are inserted to progressively enlarge the receptive field. The $k$-nearest neighbour graph is constructed with \(k=40\) for all layers, using Euclidean distance in feature space.

\section{Experiments}
We evaluate our method on three benchmarks: ModelNet40 \cite{wu20153d} for shape classification, ShapeNetPart \cite{yi2016scalable} for part segmentation, and ScanObjectNN \cite{uy2019revisiting} for real-world classification. To assess rotation invariance, we adopt three standard train/test protocols: z/z, z/SO(3), and SO(3)/SO(3), where z and SO(3) represent vertical and arbitrary rotations. For fair comparison, RIAttnConv is integrated into standard backbones without additional modifications. Extensive ablation studies further validate the contribution of each component, with more results provided in the supplementary material.

\subsection{Implementation Details}
We use stochastic gradient descent (SGD) with an initial learning rate of 0.1, decayed to 0.001 via cosine annealing \cite{loshchilov2016sgdr}. Models are trained for 300 epochs with a batch size of 32 for classification and 16 for segmentation. A dropout rate of 0.5 is applied to fully connected layers.

To ensure rotation-invariant input representations, we initialize a point-wise descriptor for each point \(p_i\) based on its relative position to the global centroid \(O\) \cite{cohen2018spherical}, including \(\|\overrightarrow{Op_i}\|_2\), \(\sin(\angle(\partial_i^1, \overrightarrow{Op_i}))\), and \(\cos(\angle(\partial_i^1, \overrightarrow{Op_i}))\), where \(\partial_i^1\) is the local reference axis. By default, we use surface normals as \(\partial_i^1\) (denoted as pc+normal). For methods using only coordinates, we replace \(\partial_i^1\) with the vector \(\overrightarrow{p_r m_r}\), from the barycenter to the point.

\begin{table}[t]
\def\arraystretch{0.8}
\centering
\scriptsize
\resizebox{0.47\textwidth}{!}{
\begin{tabular}{l c c c c}
\hline
Methods & Inputs & z/z & z/SO(3) & SO(3)/SO(3) \\
\hline
\multicolumn{5}{c}{Rotation-sensitive} \\
\hline
PointNet \cite{qi2017pointnet} & pc & 89.2 & 16.2 & 75.5 \\
PointNet++ \cite{qi2017pointnet++} & pc+n & 89.3 & 28.6 & 85.0 \\
PointCNN \cite{li2018pointcnn} & pc & 92.2 & 41.2 & 84.5 \\
DGCNN \cite{wang2019dynamic} & pc & 92.2 & 20.6 & 81.1 \\
\hline
\multicolumn{5}{c}{Rotation-robust} \\
\hline
RIConv \cite{zhang2019rotation} & pc & 86.5 & 86.4 & 86.4 \\
ClusterNet \cite{chen2019clusternet} & pc & 87.1 & 87.1 & 87.1  \\
RI-GCN \cite{kim2020rotation} & pc+n & 91.0 & 91.0 & 91.0 \\
VN-DGCNN \cite{deng2021vector} & pc & 89.5 & 89.5 & 90.2 \\
SGMNet \cite{xu2021sgmnet} & pc & 90.0 & 90.0 & 90.0 \\
OrientedMP \cite{luo2022equivariant} & pc+n & 88.4 & 88.4 & 88.4 \\
ELGANet \cite{gu2022enhanced} & pc & 90.3 & 90.3 & 90.3 \\
PaRot \cite{zhang2023parot} & pc & 90.9 & 91.0 & 90.8 \\
TetraSphere \cite{melnyk2024tetrasphere} & pc & 90.5 & 90.5 & 90.3 \\
\hline
Ours & pc & \textbf{91.8} & \textbf{91.8} & \textbf{91.8} \\
Ours & pc+n & \textbf{92.6} & \textbf{92.6} & \textbf{92.6} \\
\hline
\end{tabular}
}
\caption{Shape classification accuracy (\%) on ModelNet40 dataset. 'pc' and 'n' represent the 3D coordinates and normal vectors of input data, respectively.}
\label{tab:shape_cls}
\end{table}

\begin{table}[t]
\def\arraystretch{0.8}
\centering
\scriptsize
\resizebox{0.45\textwidth}{!}{
\begin{tabular}{l c c c}
\hline
Methods & z/SO(3) & SO(3)/SO(3) \\
\hline
\multicolumn{3}{c}{Input pc only} \\
\hline
PointNet \cite{qi2017pointnet} & 16.7 & 54.7 \\
PointNet++ \cite{qi2017pointnet++} & 15.0 & 47.4 \\
PointCNN \cite{li2018pointcnn} & 14.6 & 63.7 \\
DGCNN \cite{wang2019dynamic} & 17.7 & 71.8 \\
\hline
RIConv \cite{zhang2019rotation} & 75.3 & 75.5 \\
RI-GCN \cite{kim2020rotation} & 80.5 & 80.6 \\
VN-DGCNN \cite{deng2021vector} & 79.8 & 80.3 \\
OrientedMP \cite{luo2022equivariant} & 76.7 & 77.2 \\
LGR-Net \cite{zhao2022rotation} & 81.2 & 81.4 \\
PaRI-Conv \cite{chen2022devil} & 83.3 & 83.3 \\
PaRot \cite{zhang2023parot} & 82.1 & 82.6 \\
\hline
Ours & \textbf{84.0} & \textbf{84.0} \\
\hline
\end{tabular}
}
\caption{Real-world classification accuracy (\%) on ScanObjectNN OBJ\_BG dataset.}
\label{tab:real_cls}
\end{table}

\begin{table}[h]
\centering
\scriptsize
\resizebox{0.47\textwidth}{!}{
\begin{tabular}{l c cc cc}
\hline
\multirow{2.5}{*}{Method} & \multirow{2.5}{*}{Input} & \multicolumn{2}{c}{z/SO(3)} & \multicolumn{2}{c}{SO(3)/SO(3)} \\
\cmidrule(lr{0.0em}){3-4} \cmidrule(lr{0.0em}){5-6}
  &  & C. mIoU & I. mIoU & C. mIoU & I. mIoU \\
\hline
PointNet \cite{qi2017pointnet} & pc & 37.8 & - & 74.4 & - \\
PointNet++ \cite{qi2017pointnet++} & pc & 48.2 & - & 76.7 & - \\
DGCNN \cite{wang2019dynamic} & pc & 37.4 & - & 73.3 & - \\
\hline
RI-CNN \cite{zhang2019rotation} & pc & 75.3 & - & 75.5 & - \\
RI-GCN \cite{kim2020rotation} & pc+n & 77.2 & - & 77.3 & - \\
VN-DGCNN \cite{deng2021vector} & pc & - & 81.4 & - & 81.4 \\
LGR-Net \cite{zhao2022rotation} & pc+n & - & - & - & 82.8 \\
PaRI-Conv \cite{chen2022devil} & pc+n & - & 84.6 & - & 84.6\\
CRIN \cite{lou2023crin} & pc & 80.5 & - & 80.5 & - \\
PaRot \cite{zhang2023parot} & pc & 79.2 & - & 79.5 & - \\
RI-PCR \cite{yu2023rethinking} & pc & - & 80.3 & - & 80.4 \\
LocoTrans \cite{chen2024local} & pc & 80.1 & 84.0 & 80.0 & 83.8 \\
TetraSphere \cite{melnyk2024tetrasphere} & pc & - & 82.3 & - & 82.3 \\
RISurConv \cite{zhang2024risurconv} & pc+n & 81.5 & - & 81.5 & - \\
\hline
\textbf{Ours} & pc & \textbf{81.7} & \textbf{84.4} & \textbf{82.4} & \textbf{85.1} \\
\textbf{Ours} & pc+n & \textbf{82.9} & \textbf{85.0} & \textbf{83.8} & \textbf{85.5} \\
\hline
\end{tabular}
}
\caption{Part segmentation results on ShapeNetPart dataset. ‘C. mIoU’ stands for mIoU (\%) over all classes while ‘I. mIoU’ denotes mIoU over all instances.}
\label{tab:part_seg}
\end{table}

\subsection{3D Shape Classification}
\noindent\textbf{Dataset.}\;
We evaluate our method on the widely used ModelNet40~\cite{wu20153d} benchmark, which contains 12{,}311 CAD models from 40 categories, with 9{,}843 samples for training and 2{,}468 for testing. Following \cite{qi2017pointnet}, we randomly sample 1{,}024 points for each object and augment the training data with random scaling and translation.

\smallskip
\noindent\textbf{Results.}\;
Table~\ref{tab:shape_cls} reports classification accuracy under three rotation protocols: z/z, z/SO(3), and SO(3)/SO(3). While rotation-sensitive methods (e.g., PointCNN, DGCNN) perform well under aligned conditions, their accuracy drops sharply (up to 70\%) under unseen rotations. In contrast, RI methods exhibit stable performance across settings but often suffer from limited expressiveness without normals; for instance, RIConv and ClusterNet achieve only 86–87\% accuracy in such cases.

Our proposed RIAttnConv achieves 91.8\% accuracy using only point coordinates, surpassing several normal-dependent methods such as OrientedMP (88.4\%) and RI-GCN (91.0\%). With surface normals, performance further improves to 92.6\%, yielding a 1.6\% gain over RI-GCN and outperforming recent approaches including TetraSphere and PaRot. These results highlight the effectiveness of our attention-based design in learning pose-aware yet rotation-invariant features.

\subsection{Real-World Shape Classification}
\noindent\textbf{Dataset.}
We further evaluate our method on the ScanObjectNN dataset \cite{uy2019revisiting}, a real-world benchmark comprising 2,902 objects from 15 categories captured in cluttered and noisy environments. Following standard protocol, we use the OBJ\_BG subset, consisting of 2,890 samples with 2,312 for training and 578 for testing. As surface normals are unavailable, only raw point coordinates are used.

\smallskip
\noindent\textbf{Results.}
As shown in Table~\ref{tab:real_cls}, our method achieves the highest accuracy under both z/SO(3) and SO(3)/SO(3) settings, despite using only point coordinates. It outperforms PaRI-Conv by 0.7\% in both cases, highlighting the effectiveness of our global pose-aware attention in capturing robust local geometry under real-world noise and occlusion.

\begin{figure}[t]
\centering
    \includegraphics[width=.45\textwidth]{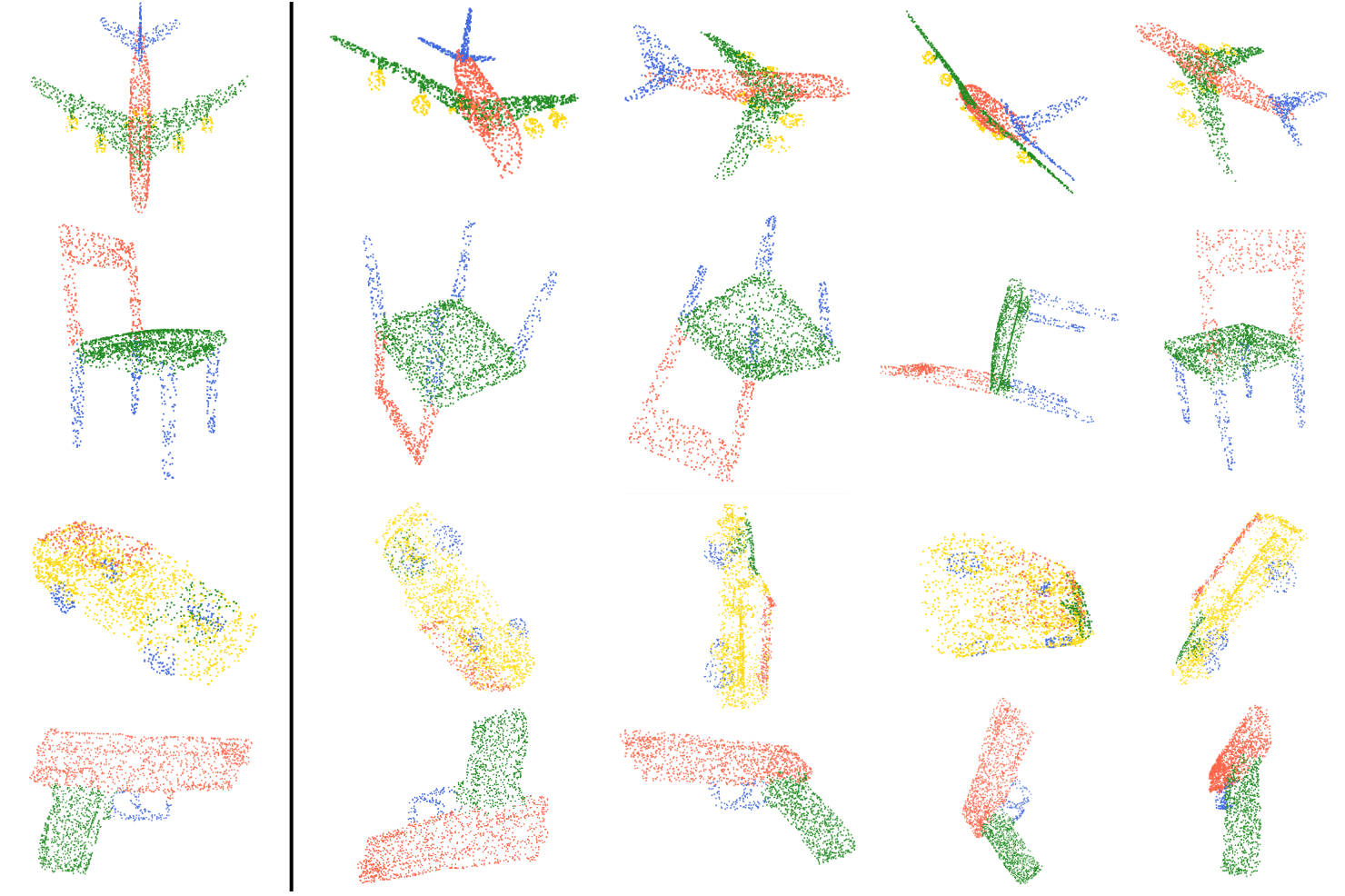}
    \caption{Visualization of part segmentation results on ShapeNetPart Dataset under z/SO(3) setting. The leftmost column is the ground truth. The testing results under arbitrary rotations are on the rest columns.}
    \label{fig_seg}
\end{figure}

\subsection{Shape Part Segmentation}
\noindent\textbf{Dataset.}
We evaluate our method on the widely adopted ShapeNetPart dataset \cite{yi2016scalable}, which contains 16,881 3D models across 16 object categories. Each object is annotated with 2 to 6 parts, yielding a total of 50 distinct part labels. Following the standard protocol in \cite{qi2017pointnet++}, we use the official train/test split and uniformly sample 2,048 points from each object for input.

\smallskip
\noindent\textbf{Results.}
As shown in Table~\ref{tab:part_seg}, our method achieves state-of-the-art performance under both z/SO(3) and SO(3)/SO(3) settings. Under the more challenging z/SO(3) setup, it achieves a class mean Intersection-over-Union (C. mIoU) of 82.9\% and an instance mean IoU (I. mIoU) of 85.0\% with normals, outperforming recent methods. Without normals, it still achieves 84.4\% and 85.1\% instance mIoU, surpassing existing RI baselines such as LocoTrans and TetraSphere, and rivaling normal-dependent approaches like PaRI-Conv and RISurConv. Furthermore, our method consistently achieves strong class-level and instance-level segmentation performance under unseen rotations. This improvement stems from the proposed SiPF and global pose-aware reference, which enable the model to disambiguate locally similar parts by leveraging global spatial context. Figure~\ref{fig_seg} visualizes representative segmentation outputs under the z/SO(3) setting, further illustrating the model’s robustness to arbitrary spatial rotations.

\begin{table}[b]
\def\arraystretch{0.8}
\scriptsize
\centering
\begin{tabular}{c c c c c}
\hline
\multirow{2.5}{*}{No.} & \multirow{2.5}{*}{RI Representation} & \multirow{2.5}{*}{Dim} & \multicolumn{2}{c}{z/SO(3)} \\
\cmidrule(lr{0.5em}){4-5}
& & & C. mIoU & I. mIoU \\

\hline
1 & PPF & 4 & 81.1 & 84.1 \\
2 & Aug. PPF & 8 & 81.8 & 84.2 \\
3 & SiPF-w/oDirection & 5 & 82.4 & 84.5 \\
4 & SiPF & 8 & \textbf{82.9} & \textbf{85.0} \\
\hline
\end{tabular}
\caption{Ablation study on the relative pose feature $\mathcal{P}_r^j$.}
\label{tab:as_rpf}
\end{table}

\subsection{Ablation Studies}
In this section, we present a series of ablation experiments to assess the effectiveness of the key components in our proposed framework. Unless otherwise stated, all experiments are performed on ShapeNetPart dataset with normal vector under the z/SO(3) setting and reported with C. mIoU (\%) and I. mIoU (\%).

\begin{table}[t]
\def\arraystretch{0.8}
\scriptsize
\centering
\begin{tabular}{c c c c c}
\hline
\multirow{2.5}{*}{RI Convolution} & \multirow{2.5}{*}{Params} & \multirow{2.5}{*}{FLOPs} & \multicolumn{2}{c}{z/SO(3)} \\
\cmidrule(lr{0.5em}){4-5}
& & & C. mIoU & I. mIoU \\

\hline
Concat. & 3.00M & 4895M & 82.1 & 84.2 \\
AdaptConv & 3.91M& 18896M & 82.0 & 84.2 \\
PaRI-Conv & 2.95M& 4574M & 82.4 & 84.5 \\
RIAttnConv & 3.01M& 4795M & \textbf{82.9} & \textbf{85.0} \\
\hline
\end{tabular}
\caption{Ablation study on the RI convolution.}
\label{tab:as_ric}
\end{table}

\begin{table}[t]
\def\arraystretch{0.8}
\scriptsize
\centering
\begin{tabular}{ccccc}
\hline
 & LocoTrans & TetraSphere & RISurConv & \textbf{Ours} \\
\hline
Params & 6.72M$^*$ & 1.31M & 4.06M & \textbf{3.01M} \\
FLOPs & 7998M$^*$ & 4092M & 9726M & \textbf{4795M} \\
C. mIoU & 80.1 & - & 81.3 & \textbf{81.7} \\
I. mIoU & 84.0 & 82.3 & - & \textbf{84.4} \\
\hline
\end{tabular}
\caption{Comparison of FLOPs, parameter count, and mIoU among different models with point coordinates only. ($^*$ indicates the results are estimated based on the classification architecture.)}
\label{tab:as_cb}
\end{table}

\smallskip
\noindent
\textbf{Effectiveness of SiPF.}\;
To assess the effectiveness of SiPF in capturing both rotation invariance and global pose cues, we compare it with the standard Point Pair Feature (PPF) and the augmented PPF (Aug. PPF) from \cite{chen2022devil}, which introduces an extra local axis to resolve PPF ambiguity. As shown in Table~\ref{tab:as_rpf}, SiPF clearly outperforms both variants, underscoring the importance of incorporating a global pose-aware reference. We also report ablations on SiPF’s directional components, confirming their contribution to enhanced robustness and discriminability.

\smallskip
\noindent
\textbf{Effectiveness of RIAttnConv.}\;
We evaluate the effectiveness of RIAttnConv by comparing it with alternative convolutional strategies, as shown in Table~\ref{tab:as_ric}. All variants employ SiPF as the RI descriptor for fair comparison. In addition to two recent convolutional baselines \cite{zhou2021adaptive, chen2022devil}, we include a variant that directly concatenates SiPF with latent features without structural modification. RIAttnConv achieves the best performance while maintaining comparable or lower complexity on parameters and FLOPs, highlighting its efficiency in utilizing SiPF.

\smallskip
\noindent
\textbf{Computational Efficiency.}\;
Table~\ref{tab:as_cb} reports the computational cost of our method compared to recent state-of-the-art approaches. While some methods improve accuracy by increasing parameters and FLOPs, our model achieves the best segmentation performance with substantially lower complexity, demonstrating superior efficiency and scalability.

\section{Conclusion}

In this paper, we proposed the Shadow-informed Pose Feature (SiPF), a novel rotation-invariant descriptor that enriches local geometric features with global pose context via a learned shadow reference. Together with the introduced RIAttnConv operator, our framework effectively distinguishes symmetric or spatially ambiguous structures and achieves strong performance across classification and segmentation benchmarks under arbitrary rotations. A current limitation is that our evaluation is conducted on object-level datasets. Scene-level assessment remains an appealing future direction, which may further highlight pose ambiguity and validate the broader applicability of our framework.




\newpage

\begin{strip}
\centering
{\LARGE \textbf{\textit{Supplementary Material for}}} \\
{\LARGE \textbf{Enhancing Rotation-Invariant 3D Learning with Global Pose Awareness and \\ Attention Mechanisms}}
\end{strip}


This supplementary material provides additional theoretical and experimental support for our proposed framework. We first present formal proofs establishing the rotation invariance of SiPF and RIAttnConv. Then, we analyze two types of geometric degeneracies where the discriminative power of SiPF may deteriorate: (1) alignment of the shadow with the local reference frame axis, and (2) coincidence between global rotation and local patch transformation. To address ambiguity, we incorporate a Bingham distribution over unit quaternions to model a shared epoch-wise global rotation. The proposed joint loss encourages alignment between task objectives and pose estimation. Additional experiments demonstrate the robustness of our approach under varying neighbourhood sizes and hyperparameter choices, and validate the effective learning of a globally consistent rotation.



\section*{A\quad Theoretical Analysis on Rotation Invariance}

We provide theoretical analysis to establish the rotation invariance of the proposed SiPF and the RIAttnConv operator. Formal proofs demonstrate that both components remain invariant under arbitrary rotation, ensuring their robustness and consistency under arbitrary pose variations.

\subsection*{A.1\quad Shadow-informed Pose Feature}
\begin{lemma}
\label{lem:sippf-ri}
Let $R_g$ be a shared global rotation and the shadow of the reference
point $p_r$ be $p_r' = p_r R_g$.  
Then $\textnormal{SiPPF}(p_r,p_r',p_j)$ is rotation invariant, i.e.
\begin{equation}
\begin{aligned}
\textnormal{SiPPF}(p_r,p_r',p_j)
      &=\textnormal{SiPPF}(p_rR,\,p_r'R,\,p_jR),
\end{aligned}
\end{equation}
where $\forall\,R\in SO(3)$ is arbitrary rotation, $p_j$ is one neighbourhood of $p_r$.
\end{lemma}

\begin{proof}
The feature is defined as
\begin{equation}\label{eq:sippf-def}
\text{SiPPF} (p_r,p_r',p_j)=
\frac{\text{PPF}(p_r,p_r')-\text{PPF}(p_j,p_r')}
     {\bigl\lVert\text{PPF}(p_r,p_r')-\text{PPF}(p_j,p_r')
     \bigr\rVert_2}.
\end{equation}
Because the original Point Pair Feature (PPF) is rotation invariant, which has been proved in \cite{deng2018ppf}, applying any $R\in\text{SO}(3)$ leaves both $\text{PPF}(p_r,p_r')$ and $\text{PPF}(p_j,p_r')$ unchanged. Consequently, their difference and its Euclidean norm are also
unchanged, so the right‑hand side of Equation \ref{eq:sippf-def} remains the same after rotation.  Hence the equality in the lemma holds and $\text{SiPPF}$ is rotation invariant.
\end{proof}

\begin{theorem}
\label{thm:sipf-ri}
Define the $\textnormal{SiPF}$ for a point pair $(p_r,p_j)$ as  
\begin{equation}
\label{eq:sipf-def}
\mathcal{P}_r^{\,j}
      =\bigl(\textnormal{PPF}(p_r,p_j),\;
             \textnormal{SiPPF}(p_r,p_r',p_j)\bigr)\in\mathbb{R}^8.
\end{equation}
Then $\mathcal{P}_r^{\,j}$ is rotation invariant.
\begin{equation}
\label{eq:sipf-ri}
\mathcal{P}_r^{\,j}
      =\bigl(\textnormal{PPF}(p_r R,\;p_j R),\;
             \textnormal{SiPPF}(p_r R,\;p_r' R,\;p_j R)\bigr),
\end{equation}
where $\forall\,R\in SO(3)$ is arbitrary rotation.
\end{theorem}

\begin{proof}
The PPF is rotation invariant, and Lemma \ref{lem:sippf-ri} has established the same property for $\text{SiPPF}$. Hence, since the concatenation of two rotation‑invariant vectors is itself rotation‑invariant, the composite feature $\mathcal{P}_r^{,j}$ defined in Equation \ref{eq:sipf-def} remains unchanged under any rotation, thereby validating Equation \ref{eq:sipf-ri}.
\end{proof}

\subsection*{A.2\quad RIAttnConv}
\begin{theorem}
The RIAttnConv Operator is defined as
\begin{equation}
\label{eq:riattnconv}
x'_r = \bigwedge_{j\in\mathcal{N}(p_r)}
      \operatorname{Attn}\left(
         \left\{
         \mathcal{W}(\mathcal{P}_r^j)\cdot x_j
         \right\}_{j=1}^k
      \right)
\end{equation}
is rotation invariant, where $\mathcal{P}_r^j$ denotes the SiPF.
\end{theorem}

\begin{proof}
Define an induced transformation $L_R$, that acts on the functions $f$ of the point cloud $P\in \mathbb{R}^{N\times3}$ as
\begin{equation}
    [L_R \circ f] (P) = f(PR)
\end{equation}
Given input feature $x_j$ is rotation invariant. Since $\mathcal{P}_r^j$ is also rotation invariant by Theorem \ref{thm:sipf-ri}, we can proceed to derive:
\begin{equation}
\begin{aligned}
L_R \circ x'_r 
&= L_R \circ
   \bigwedge_{j\in\mathcal{N}(p_r)}
   \operatorname{Attn}\left(
         \left\{
         \mathcal{W}(\mathcal{P}_r^j) \cdot x_j
         \right\}_{j=1}^k
   \right) \\
&= \bigwedge_{j\in\mathcal{N}(p_r)}
    \operatorname{Attn}\left(
         \left\{
         \mathcal{W}(L_R \circ \mathcal{P}_r^j)
         \cdot (L_R \circ x_j)
         \right\}_{j=1}^k
    \right) \\
&= \bigwedge_{j\in\mathcal{N}(p_r)}
    \operatorname{Attn}\left(
         \left\{
         \mathcal{W}(\mathcal{P}_r^j)\cdot x_j
         \right\}_{j=1}^k
    \right) = x'_r, 
\end{aligned}
\end{equation}
where the second equality follows from the invariance of $\mathcal{P}_r^j$ and $x_j$ under rotation, and the fact that $\mathcal{W}(\cdot)$ is a pointwise kernel function applied to RI input. Since both $\mathcal{W}(\mathcal{P}_r^j)$ and $x_j$ are rotation invariant, their product is also rotation invariant. Thus, the set of inputs to $\operatorname{Attn}(\cdot)$ and the aggregation operator $\bigwedge$ is rotation invariant. As these operations act only on already invariant quantities, the resulting output $x'_r$ preserves the rotation invariance.
\end{proof}

\section*{B\quad Limitations of SiPF under Geometry Degeneracy}

\subsection*{B.1\quad Degeneracy under Shadow-Axis Alignment}

We analyze the degeneracy of SiPF when the shadow point $p_r' = p_r R_g$
lies along the primary axis of the local reference frame (LRF) of $p_r$.

\begin{lemma}
\label{lem:shadow-axis}
Let $p_r' = p_r R_g$ be the shadow of $p_r$ under global rotation $R_g$, and let $\mathcal{L}(p_r) = [\partial_1(p_r),\partial_2(p_r),\partial_3(p_r)]$ be the LRF at $p_r$.
If the displacement vector $d = p_r' - p_r$ satisfies
\begin{equation}
d = \lambda\,\partial_1(p_r),\quad \lambda\in\mathbb{R}, \; \lambda \neq 0,
\end{equation}
then $\textnormal{PPF}(p_r, p_r')$ is a constant vector.
\end{lemma}

\begin{proof}
The standard PPF is given by
\begin{equation}
\begin{aligned}
\textnormal{PPF}(p_r, p_r') =&
(
\|d\|_2,\;
\cos\angle(\partial_1(p_r), d),\; \\
& \cos\angle(\partial_1(p_r'), d),\;
\cos\angle(\partial_1(p_r), \partial_1(p_r')) ).
\end{aligned}
\end{equation}
When $d = \lambda \partial_1(p_r)$, it holds that: (1) $\|d\|_2 = |\lambda|$ is constant; (2) $\angle(\partial_1(p_r), d) = 0$ or $\pi$, so $\cos(\cdot) = \pm 1$; (3) Since $R_g$ is aligned, we have $\partial_1(p_r') = \pm\,\partial_1(p_r)$, so $\cos\angle(\partial_1(p_r), \partial_1(p_r')) = \pm1$. Thus, all four elements in $\textnormal{PPF}(p_r, p_r')$ are constant under this geometric condition.
\end{proof}

\begin{theorem}
\label{thm:sipf-degen}
If $p_r' - p_r$ is colinear with the primary axis $\partial_1(p_r)$ of the LRF, then the $\textnormal{SiPF}$ $\mathcal{P}_r^{\,j}$ reduces to a purely $\textnormal{PPF}(p_r, p_j)$ representation.
\end{theorem}

\begin{proof}
Recall the definition:
\begin{equation}
\textnormal{SiPPF}(p_r, p_r', p_j) =
\frac{
\textnormal{PPF}(p_r, p_r') - \textnormal{PPF}(p_j, p_r')
}{
\left\|
\textnormal{PPF}(p_r, p_r') - \textnormal{PPF}(p_j, p_r')
\right\|_2}.
\end{equation}
From Lemma~\ref{lem:shadow-axis}, the term $\textnormal{PPF}(p_r, p_r')$ is constant,
so the entire numerator depends only on $\textnormal{PPF}(p_j, p_r')$.
Therefore,
\begin{equation}
\textnormal{SiPPF}(p_r, p_r', p_j)
\approx
f\left(\textnormal{PPF}(p_j, p_r')\right),
\end{equation}
where $f(\cdot)$ denotes a fixed affine transformation plus normalization.
This implies that the shadow brings no additional information, and SiPF becomes
\begin{equation}
\begin{aligned}
\mathcal{P}_r^{\,j}
=
&\left(
\textnormal{PPF}(p_r, p_j),\;
\textnormal{SiPPF}(p_r, p_r', p_j)
\right) \\
&\approx
\left(
\textnormal{PPF}(p_r, p_j),\;
\textnormal{PPF}(p_j, p_r')
\right),
\end{aligned}
\end{equation}
Note that although $\textnormal{PPF}(p_j, p_r)$ and $\textnormal{PPF}(p_j, p_r')$ are generally distinct, in this degenerate configuration, $p_r$ and $p_r'$ share the same LRF and spatial direction, making them equivalent from the perspective of pairwise pose encoding. Therefore, $\textnormal{PPF}(p_j, p_r')$ can be seen as a surrogate for $\textnormal{PPF}(p_j, p_r)$, leading to a degradation of SiPF's capability in encoding local geometric variations.
\end{proof}

\begin{figure}[h]
    \centering
    \includegraphics[width=.45\textwidth]{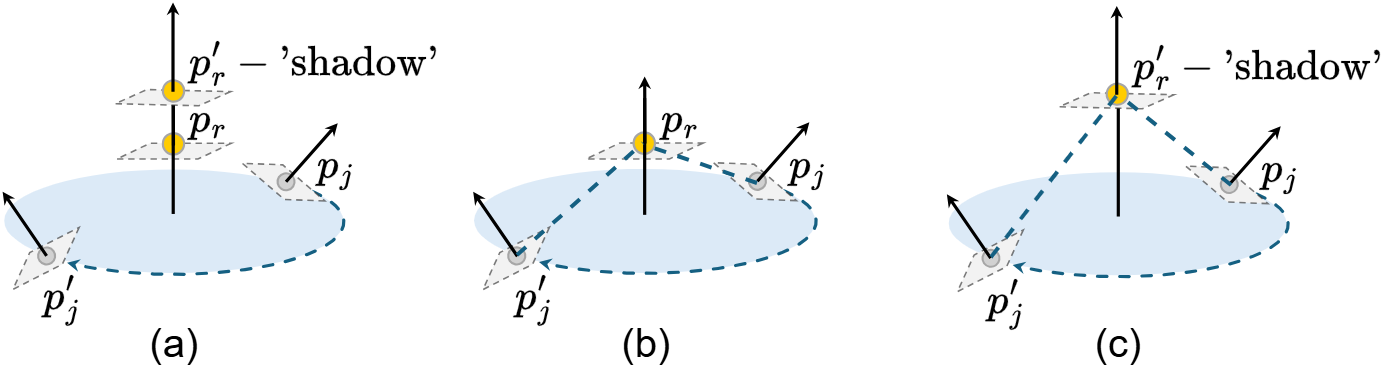}
    \caption{Illustration of the \emph{Degeneracy under Shadow–Axis Alignment}. 
    When the shadow point $p_r'$ lies along the primary LRF axis $\partial_1(p_r)$, the pairwise geometric descriptor $\textnormal{PPF}(p_r, p_r')$ becomes constant, 
    causing SiPPF to degenerate into standard PPF, which is known to be ambiguous to azimuthal variations around $\partial_1(p_r)$.
    \textbf{(a)} Reference point $p_r$ and its shadow $p_r'$ are vertically aligned, and neighbours $p_j$ and $p_j'$ lie on the circular ring centered at $p_r$.
    \textbf{(b)} A different neighbour $p_j$ with rotated azimuth but the same distance and normal still yields the same $\textnormal{PPF}(p_r,p_j)$.
    \textbf{(c)} Due to the constancy of $\textnormal{PPF}(p_r,p_r')$, the SiPPF representation collapses to $\textnormal{PPF}(p_r,p_j)$, 
    thereby losing discriminability among neighbours with different angular positions around $\partial_1(p_r)$.
    }
    \label{fig:degeneracy_case1}
\end{figure}

\subsection*{B.2\quad Degeneracy under Shadow-Local Coincidence}

We analyze the failure case when the global rotation $R_g$ used to generate the shadow $p_r' = p_r R_g$ coincides with the local patch rotation $R_j$ used in patch-swapping transformation $T(\Gamma(p_r))$.

\begin{lemma}
\label{lem:shadow-local}
If the global shadow rotation satisfies $R_g = R_j$, then for any point $p_j$ in the receptive field $\mathcal{N}(p_r)$, the SiPPF feature
\begin{equation}
\textnormal{SiPPF}(p_r, p_r', p_j) =
\frac{
\textnormal{PPF}(p_r, p_r') - \textnormal{PPF}(p_j, p_r')
}{
\left\|
\textnormal{PPF}(p_r, p_r') - \textnormal{PPF}(p_j, p_r')
\right\|_2}
\end{equation}
is invariant to replacing $\Omega(p_j)$ with its rotated counterpart $\Omega(p_j) R_j$.
\end{lemma}

\begin{proof}
Let $p_j^R = p_j R_j$ be the rotated version of $p_j$ under patch-swapping transformation.
Since $R_j = R_g$, we have:
\begin{equation}
p_r' = p_r R_g = p_r R_j.
\end{equation}
Then the rotated shadow becomes:
\begin{equation}
p_r'^R = (p_r R_j) R_g = (p_r R_j) R_j = p_r' R_j.
\end{equation}

This means the shadow of $p_r^R$ is exactly the rotated version of $p_r'$:
\begin{equation}
p_r'^R = p_r' R_j.
\end{equation}
The equality \emph{does not hold in general}, but is specific to the degenerate case $R_g = R_j$. From the definition of $\textnormal{PPF}$, which is rotation invariant, we have:
\begin{equation}
\begin{aligned}
& \textnormal{PPF}(p_j, p_r') = \textnormal{PPF}(p_j^R,\, p_r'^R), \\
& \textnormal{PPF}(p_r, p_r') = \textnormal{PPF}(p_r^R,\, p_r'^R).
\end{aligned}
\end{equation}
Now plug into the SiPPF definition:
\begin{align}
\textnormal{SiPPF}(p_r, p_r', p_j)
&= \frac{
\textnormal{PPF}(p_r, p_r') - \textnormal{PPF}(p_j, p_r')
}{
\left\| \textnormal{PPF}(p_r, p_r') - \textnormal{PPF}(p_j, p_r') \right\|_2} \\
&= \frac{
\textnormal{PPF}(p_r^R, p_r'^R) - \textnormal{PPF}(p_j^R, p_r'^R)
}{
\left\| \textnormal{PPF}(p_r^R, p_r'^R) - \textnormal{PPF}(p_j^R, p_r'^R) \right\|_2} \\
&= \textnormal{SiPPF}(p_r^R, p_r'^R, p_j^R)
\end{align}

This implies that if we replace a local patch $\Omega(p_j)$ with its symmetric counterpart $\Omega(p_j)R_j$, and simultaneously apply $R_j$ to $p_r$ and its shadow $p_r'$, the resulting SiPPF remains unchanged. Hence, SiPPF is invariant to patch-level symmetric transformation when $R_g = R_j$, and fails to distinguish symmetric regions such as mirrored wings or legs.
\end{proof}


\begin{theorem}
If the global rotation $R_g$ coincides with the local patch rotation $R_j$,  
then the SiPF feature $\mathcal{P}_r^j$ cannot distinguish $\Omega(p_j)$ from its rotated variant $\Omega(p_j)R_j$.
\end{theorem}

\begin{proof}
From Lemma 1, $\textnormal{SiPPF}$ is rotation invariant. This means that if a local patch $\Omega(p_j)$ is replaced with its symmetric counterpart $\Omega(p_j) R_j$,  
and the reference point and shadow are rotated accordingly, the resulting SiPPF remains unchanged. Consequently, the full SiPF:
\begin{equation}
\mathcal{P}_r^j = \left( \textnormal{PPF}(p_r, p_j),\; \textnormal{SiPPF}(p_r, p_r', p_j) \right)
\end{equation}
also remains invariant under $p_j = p_j R_j$. Hence, SiPF fails to distinguish between symmetric patches when $R_g = R_j$, leading to a degeneration into a representation lacking global discriminability.
\end{proof}

\begin{figure}[t]
    \centering
    \includegraphics[width=.45\textwidth]{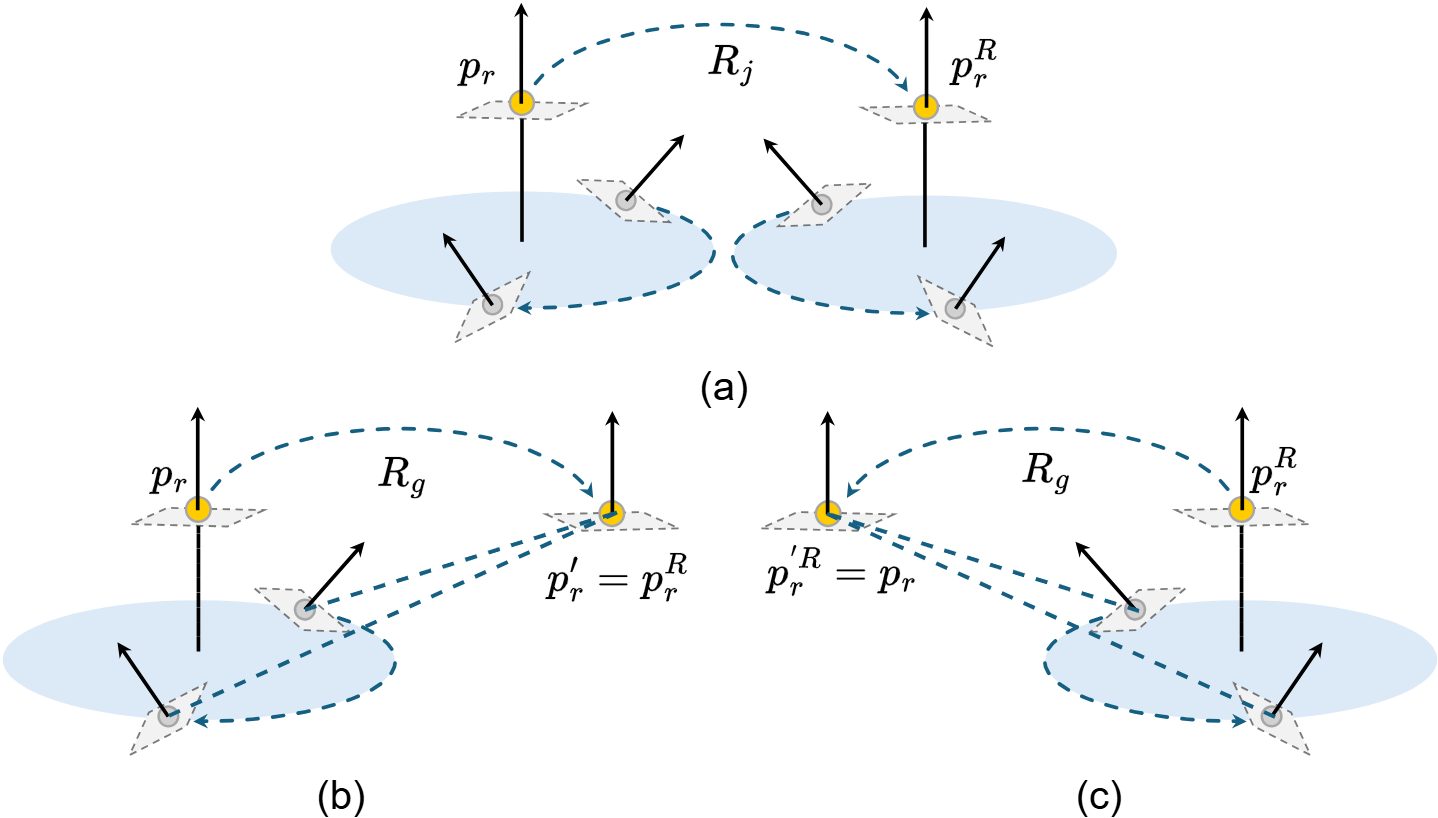}
    \caption{
    Illustration of the \emph{Degeneracy under Shadow–Local Coincidence}. When the global rotation \( R_g \) coincides with the patch-swapping transformation \( R_j \), the SiPPF descriptor becomes insensitive to residual in-plane rotations of \( p_j \), failing to distinguish the local receptive field \( \Omega(p_j) \) from its rotated counterpart \( \Omega(p_j) R_j \).
    \textbf{(a)} The local reference point \( p_r \) is rotated to \( p_r^R \) through the local transformation \( R_j \), aligning the LRF with the receptive field.
    \textbf{(b)} The global rotation \( R_g \) is equal to \( R_j \), resulting in the shadow \( p_r' = p_r^R \).
    \textbf{(c)} The inverse of \( R_g \) is applied to \( p_r^R \), resulting in \( p_r^{'R} = p_r \). This makes the SiPF invariant to \( R_j \), hence losing its ability to distinguish between the original receptive field \( \Omega(p_j) \) and the rotated variant \( \Omega(p_j) R_j \).
    }
    \label{fig:degeneracy_case2}
\end{figure}

\section*{C\quad Bingham distribution on the hypersphere of unit quaternions}

\subsection*{C.1\quad Bingham distribution}
The probability density function of \textit{Bingham} distribution \cite{bingham1974antipodally, glover2014quaternion} for a unit quaternion $\mathbf{q} \in \mathbb{S}^{3}$ with parameter matrices $\mathbf{V}$ and $\boldsymbol{\Lambda}$ is:
\begin{equation}
\label{eq_bin}
\mathcal{B}(\mathbf{q}\mid\mathbf{V}, \boldsymbol{\Lambda})=\frac{1}{F(\boldsymbol{\Lambda})} \exp \left( \mathbf{q}^{T} \mathbf{V} \boldsymbol{\Lambda} \mathbf{V}^{T} \mathbf{q} \right),
\end{equation}
where $\boldsymbol{\Lambda} = \text{diag}(\lambda_1,\lambda_2, \lambda_3, 0) $ is a diagonal matrix with negative parameters $(\lambda_1 \leq \lambda_2 \leq \lambda_3 < 0)$, and the columns of the matrix $\mathbf{V}\in\mathbb{R}^{4\times 4}$ are orthogonal unit vectors ${\mathbf{v}}$. $F(\boldsymbol{\Lambda})$ is the normalizing constant.

\subsection*{C.2\quad Entropy of Bingham distribution}

\begin{theorem}
The \textbf{entropy} of a Bingham distribution with PDF $f$ is given by:
\begin{equation}
\label{eq_entropy_bhms}
h(f) = \log F - \boldsymbol{\Lambda} \cdot \frac{\nabla F}{F}.
\end{equation}
\end{theorem}

\begin{proof}
Denote $C = \mathbf{V} \boldsymbol{\Lambda} \mathbf{V}^{T}$ for a more succinct form.
\begin{align*}
h(f) &= - \int_{\mathbf{q} \in S^3} f(\mathbf{q}) \log f(\mathbf{q}) \\
     &= - \int_{\mathbf{q} \in S^3} \frac{1}{F} \exp(\mathbf{q}^T C \mathbf{q}) \left(\mathbf{q}^T C \mathbf{q} - \log F\right) \\
     &= \log F - \frac{1}{F} \int_{\mathbf{q} \in S^3} \mathbf{q}^T C \mathbf{q} \exp(\mathbf{q}^T C \mathbf{q}).
\end{align*}
Using $g(\boldsymbol{\Lambda})$ to represent the hyperspherical integral $\int_{\mathbf{q} \in S^3} \mathbf{q}^T C \mathbf{q} \exp(\mathbf{q}^T C \mathbf{q})$,
\begin{align*}
g(\boldsymbol{\Lambda}) &= \int_{\mathbf{q} \in S^3} \sum_{i=1}^{3} \lambda_i (\mathbf{v}_i^T \mathbf{q})^2 \exp\left(\sum_{j=1}^{3} \lambda_j (\mathbf{v}_j^T \mathbf{q})^2\right) \\
           &= \sum_{i=1}^{3} \lambda_i \frac{\partial F}{\partial \lambda_i} = \boldsymbol{\Lambda} \cdot \nabla F.
\end{align*}
Thus, the entropy is $\log F - \boldsymbol{\Lambda} \cdot \frac{\nabla F}{F}$ as $\mathcal{L}_{\text{bingham}}$ in the main paper.
\end{proof}

\subsection*{C.3\quad $\text{Construction of } \mathbf{V}$ and $\boldsymbol{\Lambda}$}
We randomly initialize a 7-d vector $(\mathbf{z}_1, \mathbf{z}_2)$ where the first 4-d vector $\mathbf{z}_1$ are normalized as a quaternion first and then following \textit{Birdal Strategy} \cite{birdal2018bayesian} to construct the parameter metric $\mathbf{V}$ as:
\begin{equation}
    \mathbf{V}(\mathbf{z}_1) \triangleq
\begin{bmatrix}
z_{11} & -z_{12} & -z_{13} & z_{14} \\
z_{12} & z_{11} & z_{14} & z_{13} \\
z_{13} & -z_{14} & z_{11} & -z_{12} \\
z_{14} & z_{13} & -z_{12} & -z_{11}
\end{bmatrix}.
\end{equation}
The last 3-d vector $\mathbf{z}_2$ is first activated by the \textit{softplus} function $\phi$, and then transformed into a parameter matrix $\boldsymbol{\Lambda}$ with decreasing diagonal values obtained through the negative accumulative sum:

\begin{equation}
\begin{aligned}
\lambda_1 &= -\phi(z_{21}) - \phi(z_{22}) - \phi(z_{23}) \\
\lambda_2 &= -\phi(z_{21}) - \phi(z_{22}) \\
\lambda_3 &= -\phi(z_{21}).
\end{aligned}
\end{equation}

\subsection*{C.4\quad Acceptance Rejection Sampling for Bingham distribution}

Kent and Ganeiber \cite{kent2013new} demonstrated that samples from the Bingham distribution can be generated by employing the ACG distribution as an envelope density within an acceptance-rejection sampling framework. Building on this approach, the following algorithm can be applied to simulate a Bingham-based latent space with the parameter matrix $\mathbf{A} = \mathbf{V} \boldsymbol{\Lambda} \mathbf{V}^{T}$:
\begin{enumerate}
    \item Set acceptance list $\mathbf{L}$, $\underline{\boldsymbol{\Psi}}^{-1}=\mathbf{I} +\frac{2}{b}\underline{\mathbf{A}}$, and assume that there exists a constant $\mathbf{M}^* \geq \text{upper} \{ \frac{f^*(\mathbf{q}_n)}{g^*(\mathbf{q}_n)} \} $;

    \item Draw $\mathbf{u}$ from $\textit{Uniform}(0,1)$ and candidate values $\mathcal{Q}$ from the ACG distribution on spheres with parameter matrices $\underline{\boldsymbol{\Psi}}$;

    \item Update acceptance list $\mathbf{L}$, where $\mathbf{u} < \frac{f^{*}(\mathbf{q}_n; \mathbf{A}_n)}{\mathbf{M}^{*}g^{*}(\mathbf{q}_n; \boldsymbol{\Psi}_n)}$, to accept $\mathcal{Q}$. Go to Step 1 if not $\mathbf{L}\text{.all()}$. 
\end{enumerate}
Here, $f^{*}(\mathbf{q}_n; \mathbf{A}_n) = \text{exp}(-\mathbf{q}_{n}^{T}\mathbf{A}_n\mathbf{q}_n)$, and $g^{*}(\mathbf{q}_n; \boldsymbol{\Psi}_n)=(\mathbf{q}_{n}^{T}\boldsymbol{\Psi}_{n}^{-1}\mathbf{q}_n)^{-2}$. $\underline{\boldsymbol{\Psi}}$ and $\underline{\mathbf{A}}$ are the set of parameter matrices of the ACG and Bingham distribution. Empirically, setting $b=1$ as default works well in many situations. The introduction of the acceptance list $\mathbf{L}$ has significantly accelerated the speed of parallel sampling and optimization across the entire network.

\renewcommand{\algorithmicrequire}{\textbf{Input:}}  
\renewcommand{\algorithmicensure}{\textbf{Output:}}

\begin{algorithm}[t]
\caption{Enhancing Rotation-Invariant 3D Learning with Global Pose Awareness and Attention Mechanisms}
\label{alg:epoch_bingham}
\begin{algorithmic}[1]
\REQUIRE Point cloud dataset $\mathcal{D} = \{P_i\}$, network $f_\theta$, total epochs $T$
\ENSURE Trained network parameters $\theta$
\STATE Initialize network parameters $\theta$
\STATE Initialize Bingham parameters $\mathbf{V}, \boldsymbol{\Lambda}$ 
\STATE Sample global rotation quaternion: $q_g \sim \mathcal{B}(q \mid \mathbf{V}, \boldsymbol{\Lambda})$
\STATE Convert to rotation matrix: $R_g = \mathrm{quat2mat}(q_g)$
\FOR{each epoch $t = 1, 2, \dots, T$}
    \STATE Fix $R_g$ for all mini-batches in epoch $t$
    \FOR{each mini-batch $B \subset \mathcal{D}$}
        \FOR{each reference point $p_r \in B$}
            \STATE Compute shadow point: $p_r' = p_r R_g$
            \STATE Extract SiPF for each neighbor $p_j \in \mathcal{N}(p_r)$:
            \[
            \mathcal{P}_r^j = (\textnormal{PPF}(p_r, p_j),\;\textnormal{SiPPF}(p_r', p_r, p_j))
            \]
        \ENDFOR
        \STATE Forward pass: $\hat{y} = f_\theta(B;\, R_g)$
        \STATE Compute task loss: $\mathcal{L}_{\text{task}} = \mathcal{L}(\hat{y}, y)$
        \STATE Compute total loss: 
        \[
            \mathcal{L}_{\text{total}} = 
            \mathcal{L}_{\text{task}} 
            + \delta \cdot 
            \sqrt{ 
                \left( 
                    \mathcal{L}_{\text{bingham}} - 0.1 \cdot \mathcal{L}_{\text{task}} 
                \right)^2 
            }
        \]
        \STATE Update model parameters: $\theta \leftarrow \theta - \eta \nabla_\theta \mathcal{L}_{\text{total}}$
    \ENDFOR
    \STATE Update Bingham:
    \STATE \quad Resample $q_g \sim \mathcal{B}(q \mid \mathbf{V}, \boldsymbol{\Lambda})$
    \STATE \quad Update $R_g = \mathrm{quat2mat}(q_g)$
\ENDFOR
\RETURN $\theta$
\end{algorithmic}
\end{algorithm}

\section*{D\quad Epoch-wise Shadow Locating wth Bingham Distribution}

To ensure global consistency and rotation-invariance across mini-batches during training, we introduce an \textit{epoch-wise shadow locating} strategy. Specifically, we employ a shared global rotation $R_g$ to generate shadow points $p_r' = p_r R_g$, enabling relative pose computation with consistent reference directions.

Rather than sampling $R_g$ from a uniform distribution $\mathcal{U}(\mathrm{SO}(3))$ in each iteration, we learn a task-adaptive rotation distribution using the Bingham distribution.

At the beginning of each epoch, a global rotation anchor $R_g$ is sampled via $q_g \sim \mathcal{B}(q \mid \mathbf{V}, \boldsymbol{\Lambda})$ and fixed across all mini-batches in that epoch. This consistent reference facilitates the learning of rotation-invariant representations, while still allowing dynamic adaptation across epochs. 

The Bingham parameters $(\mathbf{V}, \boldsymbol{\Lambda})$ are updated throughout training via the following total loss:
\begin{equation}
\mathcal{L}_{\text{total}} = \mathcal{L}_{\text{task}} + \delta \cdot 
\sqrt{ \left( \mathcal{L}_{\text{bingham}} - 0.1 \cdot \mathcal{L}_{\text{task}} \right)^2 },
\label{eq:totloss}
\end{equation}
which encourages the Bingham mode (i.e., the most likely rotation) to align with the task optimization direction.

The complete training procedure is summarized in Algorithm~\ref{alg:epoch_bingham}. This formulation balances local patch-level pose invariance and global discriminability, and helps address ambiguities induced by symmetric shapes or local geometric degeneracies.

\section*{E\quad Additional Experiments}

\subsection*{E.1\quad Robustness to the Number of Neighbours}

The number of neighbours $k$ used in $k$NN-based local feature grouping plays a crucial role in determining the receptive field of each point during feature extraction. To assess the robustness of our model to varying neighbourhood sizes, we conduct experiments on ModelNet40 under the z/SO(3) setting. As shown in Table~\ref{tab:neighbors_accuracy}, our method maintains consistently high classification accuracy across a wide range of $k$, demonstrating strong robustness to neighbourhood configuration. Notably, performance is optimal at $k=20$, while both smaller ($k=10$) and larger ($k=40$) values still yield competitive results.

\begin{table}[h]
\centering
\begin{tabular}{c c c c}
\hline
Number of neighbors & $k$=10 & $k$=20 & $k$=40 \\
\hline
Ours (pc) & 91.5 & 91.8 & 91.7\\
Ours (pc+n) & 91.7 & 92.6 & 92.6 \\
\hline
\end{tabular}
\caption{Classification accuracy (\%) under different neighbor size $k$ on ModelNet40 z/SO(3) setting with normals.}
\label{tab:neighbors_accuracy}
\end{table}

\begin{figure*}[t]
    \centering
    \includegraphics[width=.9\textwidth]{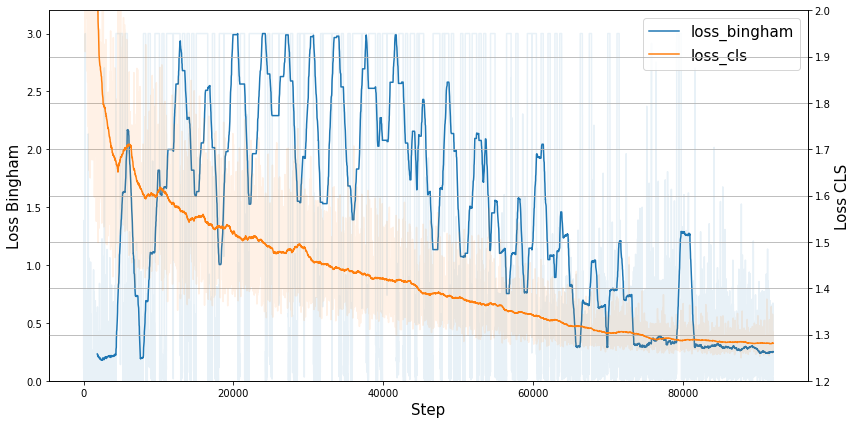}
    \caption{The Bingham loss $\mathcal{L}_{\text{bingham}}$ and classification loss $\mathcal{L}_{\text{cls}}$ in the training of ModelNet40 z/SO(3) setting. $\mathcal{L}_{\text{bingham}}$ is optimized with $\mathcal{L}_{\text{cls}}$ through our joint loss design, allowing to effectively learn a global consistent rotation $R_g$.}
    \label{fig:bingham_loss_curve}
\end{figure*}

\subsection*{E.2\quad Performance Under Different Hyperparameter $\delta$}

We evaluate the sensitivity of our method to the hyperparameter $\delta$, which balances the contribution between the task loss $\mathcal{L}_{\text{task}}$ and the pose regularization loss $\mathcal{L}_{\text{bingham}}$ in Equation \ref{eq:totloss}. As shown in Table~\ref{tab:delta_acc}, our model demonstrates stable classification performance across a wide range of $\delta$ values from $0.6$ to $1.0$. The optimal performance is achieved when $\delta = 0.8$, and the accuracy remains competitive with larger or smaller values, indicating the robustness of our formulation to the choice of this hyperparameter.

\begin{table}[h]
\centering
\begin{tabular}{c c c c c c}
\hline
$\delta$ & 0.6 & 0.7 & 0.8 & 0.9 & 1.0 \\
\hline
Ours (pc) & 91.7 & 91.7 & 91.8 & 91.5 & 91.5\\
Ours (pc+n) & 92.3 & 92.5 & 92.6 & 92.6 & 92.0 \\
\hline
\end{tabular}
\caption{Classification accuracy (\%) under different hyperparameter $\delta$ on ModelNet40 z/SO(3) setting with normals.}
\label{tab:delta_acc}
\end{table}

\subsection*{E.3\quad Learned Bingham Distribution for $R_g$}

We investigate the training behaviour of the Bingham distribution used to model the shared global rotation $R_g$. Figure \ref{fig:bingham_loss_curve} shows the evolution of the Bingham loss $\mathcal{L}_{\text{bingham}}$ and classification loss $\mathcal{L}_{\text{cls}}$ throughout training. Our total loss is formulated as Equation \ref{eq:totloss} which encourages $\mathcal{L}_{\text{bingham}}$ to track the classification loss $\mathcal{L}_{\text{task}}$. This design ensures that the optimization of the Bingham distribution aligns with the main task objective. As $\mathcal{L}_{\text{task}}$ decreases, $\mathcal{L}_{\text{bingham}}$ is gradually minimized as well, guiding the model to learn an optimal and consistent global rotation anchor $R_g$ without requiring additional supervision.


\newpage
\section{Acknowledgments}
The completion of this work was supported in part by Natural Sciences and Engineering Research Council of Canada (NSERC), the National Natural Science Foundation of China (62276106, 62576143), Guangdong Basic and Applied Basic Research Foundation (2024A1515011767), Guangdong Provincial Key Laboratory IRADS (2022B1212010006), and Guangdong Higher Education Upgrading Plan (2021-2025) of “Rushing to the Top, Making Up Shortcomings and Strengthening Special Features” (2024KTSCX222, 2024GXJK695).

\bibliography{arxiv}

\begin{thebibliography}{45}
\providecommand{\natexlab}[1]{#1}

\bibitem[{Afham et~al.(2022)Afham, Dissanayake, Dissanayake, Dharmasiri,
  Thilakarathna, and Rodrigo}]{afham2022crosspoint}
Afham, M.; Dissanayake, I.; Dissanayake, D.; Dharmasiri, A.; Thilakarathna, K.;
  and Rodrigo, R. 2022.
\newblock Crosspoint: Self-supervised cross-modal contrastive learning for 3d
  point cloud understanding.
\newblock In \emph{Proceedings of the IEEE/CVF conference on computer vision
  and pattern recognition}, 9902--9912.

\bibitem[{Bingham(1974)}]{bingham1974antipodally}
Bingham, C. 1974.
\newblock An antipodally symmetric distribution on the sphere.
\newblock \emph{The Annals of Statistics}, 1201--1225.

\bibitem[{Birdal et~al.(2018)Birdal, Simsekli, Eken, and
  Ilic}]{birdal2018bayesian}
Birdal, T.; Simsekli, U.; Eken, M.~O.; and Ilic, S. 2018.
\newblock Bayesian pose graph optimization via bingham distributions and
  tempered geodesic mcmc.
\newblock \emph{Advances in neural information processing systems}, 31.

\bibitem[{Chen et~al.(2019)Chen, Li, Xu, Chen, Wang, and
  Lin}]{chen2019clusternet}
Chen, C.; Li, G.; Xu, R.; Chen, T.; Wang, M.; and Lin, L. 2019.
\newblock Clusternet: Deep hierarchical cluster network with rigorously
  rotation-invariant representation for point cloud analysis.
\newblock In \emph{Proceedings of the IEEE/CVF conference on computer vision
  and pattern recognition}, 4994--5002.

\bibitem[{Chen and Cong(2022)}]{chen2022devil}
Chen, R.; and Cong, Y. 2022.
\newblock The devil is in the pose: Ambiguity-free 3d rotation-invariant
  learning via pose-aware convolution.
\newblock In \emph{Proceedings of the IEEE/CVF Conference on Computer Vision
  and Pattern Recognition}, 7472--7481.

\bibitem[{Chen et~al.(2024)Chen, Duan, Zhao, Ding, and Tao}]{chen2024local}
Chen, Y.; Duan, L.; Zhao, S.; Ding, C.; and Tao, D. 2024.
\newblock Local-consistent transformation learning for rotation-invariant point
  cloud analysis.
\newblock In \emph{Proceedings of the IEEE/CVF Conference on Computer Vision
  and Pattern Recognition}, 5418--5427.

\bibitem[{Cohen et~al.(2018)Cohen, Geiger, K{\"o}hler, and
  Welling}]{cohen2018spherical}
Cohen, T.~S.; Geiger, M.; K{\"o}hler, J.; and Welling, M. 2018.
\newblock Spherical cnns.
\newblock \emph{arXiv preprint arXiv:1801.10130}.

\bibitem[{Deng et~al.(2021)Deng, Litany, Duan, Poulenard, Tagliasacchi, and
  Guibas}]{deng2021vector}
Deng, C.; Litany, O.; Duan, Y.; Poulenard, A.; Tagliasacchi, A.; and Guibas,
  L.~J. 2021.
\newblock Vector neurons: A general framework for so (3)-equivariant networks.
\newblock In \emph{Proceedings of the IEEE/CVF International Conference on
  Computer Vision}, 12200--12209.

\bibitem[{Deng, Birdal, and Ilic(2018)}]{deng2018ppf}
Deng, H.; Birdal, T.; and Ilic, S. 2018.
\newblock Ppf-foldnet: Unsupervised learning of rotation invariant 3d local
  descriptors.
\newblock In \emph{Proceedings of the European conference on computer vision
  (ECCV)}, 602--618.

\bibitem[{Drost et~al.(2010)Drost, Ulrich, Navab, and Ilic}]{drost2010model}
Drost, B.; Ulrich, M.; Navab, N.; and Ilic, S. 2010.
\newblock Model globally, match locally: Efficient and robust 3D object
  recognition.
\newblock In \emph{2010 IEEE computer society conference on computer vision and
  pattern recognition}, 998--1005. Ieee.

\bibitem[{Duan et~al.(2023)Duan, Zhao, Xue, Gong, Xia, and
  Tao}]{duan2023condaformer}
Duan, L.; Zhao, S.; Xue, N.; Gong, M.; Xia, G.-S.; and Tao, D. 2023.
\newblock Condaformer: Disassembled transformer with local structure
  enhancement for 3d point cloud understanding.
\newblock \emph{Advances in Neural Information Processing Systems}, 36:
  23886--23901.

\bibitem[{Fuchs et~al.(2020)Fuchs, Worrall, Fischer, and Welling}]{fuchs2020se}
Fuchs, F.; Worrall, D.; Fischer, V.; and Welling, M. 2020.
\newblock Se (3)-transformers: 3d roto-translation equivariant attention
  networks.
\newblock \emph{Advances in neural information processing systems}, 33:
  1970--1981.

\bibitem[{Glover(2014)}]{glover2014quaternion}
Glover, J.~M. 2014.
\newblock \emph{The quaternion Bingham distribution, 3D object detection, and
  dynamic manipulation}.
\newblock Ph.D. thesis, Massachusetts Institute of Technology.

\bibitem[{Gu et~al.(2022)Gu, Wu, Li, Kang, Ng, and Wang}]{gu2022enhanced}
Gu, R.; Wu, Q.; Li, Y.; Kang, W.; Ng, W.~W.; and Wang, Z. 2022.
\newblock Enhanced local and global learning for rotation-invariant point cloud
  representation.
\newblock \emph{IEEE MultiMedia}, 29(4): 24--37.

\bibitem[{Guo et~al.(2020)Guo, Wang, Hu, Liu, Liu, and Bennamoun}]{guo2020deep}
Guo, Y.; Wang, H.; Hu, Q.; Liu, H.; Liu, L.; and Bennamoun, M. 2020.
\newblock Deep learning for 3d point clouds: A survey.
\newblock \emph{IEEE transactions on pattern analysis and machine
  intelligence}, 43(12): 4338--4364.

\bibitem[{Jin et~al.(2023)Jin, Hayat, Yang, Guo, and Lei}]{jin2023context}
Jin, Z.; Hayat, M.; Yang, Y.; Guo, Y.; and Lei, Y. 2023.
\newblock Context-aware alignment and mutual masking for 3d-language
  pre-training.
\newblock In \emph{Proceedings of the IEEE/CVF Conference on Computer Vision
  and Pattern Recognition}, 10984--10994.

\bibitem[{Jing et~al.(2020)Jing, Eismann, Suriana, Townshend, and
  Dror}]{jing2020learning}
Jing, B.; Eismann, S.; Suriana, P.; Townshend, R.~J.; and Dror, R. 2020.
\newblock Learning from protein structure with geometric vector perceptrons.
\newblock \emph{arXiv preprint arXiv:2009.01411}.

\bibitem[{Kent, Ganeiber, and Mardia(2013)}]{kent2013new}
Kent, J.~T.; Ganeiber, A.~M.; and Mardia, K.~V. 2013.
\newblock A new method to simulate the Bingham and related distributions in
  directional data analysis with applications.
\newblock \emph{arXiv preprint arXiv:1310.8110}.

\bibitem[{Kim, Park, and Han(2020)}]{kim2020rotation}
Kim, S.; Park, J.; and Han, B. 2020.
\newblock Rotation-invariant local-to-global representation learning for 3d
  point cloud.
\newblock \emph{Advances in Neural Information Processing Systems}, 33:
  8174--8185.

\bibitem[{Li et~al.(2021{\natexlab{a}})Li, Fujiwara, Okura, and
  Matsushita}]{li2021closer}
Li, F.; Fujiwara, K.; Okura, F.; and Matsushita, Y. 2021{\natexlab{a}}.
\newblock A closer look at rotation-invariant deep point cloud analysis.
\newblock In \emph{Proceedings of the IEEE/CVF International Conference on
  Computer Vision}, 16218--16227.

\bibitem[{Li et~al.(2021{\natexlab{b}})Li, Li, Chen, Fu, Cohen-Or, and
  Heng}]{li2021rotation}
Li, X.; Li, R.; Chen, G.; Fu, C.-W.; Cohen-Or, D.; and Heng, P.-A.
  2021{\natexlab{b}}.
\newblock A rotation-invariant framework for deep point cloud analysis.
\newblock \emph{IEEE transactions on visualization and computer graphics},
  28(12): 4503--4514.

\bibitem[{Li et~al.(2018)Li, Bu, Sun, Wu, Di, and Chen}]{li2018pointcnn}
Li, Y.; Bu, R.; Sun, M.; Wu, W.; Di, X.; and Chen, B. 2018.
\newblock Pointcnn: Convolution on x-transformed points.
\newblock \emph{Advances in neural information processing systems}, 31.

\bibitem[{Liu et~al.(2019)Liu, Fan, Xiang, and Pan}]{liu2019relation}
Liu, Y.; Fan, B.; Xiang, S.; and Pan, C. 2019.
\newblock Relation-shape convolutional neural network for point cloud analysis.
\newblock In \emph{Proceedings of the IEEE/CVF conference on computer vision
  and pattern recognition}, 8895--8904.

\bibitem[{Loshchilov and Hutter(2016)}]{loshchilov2016sgdr}
Loshchilov, I.; and Hutter, F. 2016.
\newblock Sgdr: Stochastic gradient descent with warm restarts.
\newblock \emph{arXiv preprint arXiv:1608.03983}.

\bibitem[{Lou et~al.(2023)Lou, Ye, You, Jiang, Lu, Wang, Ma, and
  Lu}]{lou2023crin}
Lou, Y.; Ye, Z.; You, Y.; Jiang, N.; Lu, J.; Wang, W.; Ma, L.; and Lu, C. 2023.
\newblock CRIN: rotation-invariant point cloud analysis and rotation estimation
  via centrifugal reference frame.
\newblock In \emph{Proceedings of the AAAI Conference on Artificial
  Intelligence}, volume~37, 1817--1825.

\bibitem[{Luo et~al.(2022)Luo, Li, Guan, Su, Cheng, Peng, and
  Ma}]{luo2022equivariant}
Luo, S.; Li, J.; Guan, J.; Su, Y.; Cheng, C.; Peng, J.; and Ma, J. 2022.
\newblock Equivariant point cloud analysis via learning orientations for
  message passing.
\newblock In \emph{Proceedings of the IEEE/CVF Conference on Computer Vision
  and Pattern Recognition}, 18932--18941.

\bibitem[{Melnyk et~al.(2024)Melnyk, Robinson, Felsberg, and
  Wadenb{\"a}ck}]{melnyk2024tetrasphere}
Melnyk, P.; Robinson, A.; Felsberg, M.; and Wadenb{\"a}ck, M. 2024.
\newblock Tetrasphere: A neural descriptor for o (3)-invariant point cloud
  analysis.
\newblock In \emph{Proceedings of the IEEE/CVF Conference on Computer Vision
  and Pattern Recognition}, 5620--5630.

\bibitem[{Poulenard and Guibas(2021)}]{poulenard2021functional}
Poulenard, A.; and Guibas, L.~J. 2021.
\newblock A functional approach to rotation equivariant non-linearities for
  Tensor Field Networks.
\newblock In \emph{Proceedings of the IEEE/CVF Conference on Computer Vision
  and Pattern Recognition}, 13174--13183.

\bibitem[{Qi et~al.(2017{\natexlab{a}})Qi, Su, Mo, and Guibas}]{qi2017pointnet}
Qi, C.~R.; Su, H.; Mo, K.; and Guibas, L.~J. 2017{\natexlab{a}}.
\newblock Pointnet: Deep learning on point sets for 3d classification and
  segmentation.
\newblock In \emph{Proceedings of the IEEE conference on computer vision and
  pattern recognition}, 652--660.

\bibitem[{Qi et~al.(2017{\natexlab{b}})Qi, Yi, Su, and
  Guibas}]{qi2017pointnet++}
Qi, C.~R.; Yi, L.; Su, H.; and Guibas, L.~J. 2017{\natexlab{b}}.
\newblock Pointnet++: Deep hierarchical feature learning on point sets in a
  metric space.
\newblock \emph{Advances in neural information processing systems}, 30.

\bibitem[{Uy et~al.(2019)Uy, Pham, Hua, Nguyen, and Yeung}]{uy2019revisiting}
Uy, M.~A.; Pham, Q.-H.; Hua, B.-S.; Nguyen, T.; and Yeung, S.-K. 2019.
\newblock Revisiting point cloud classification: A new benchmark dataset and
  classification model on real-world data.
\newblock In \emph{Proceedings of the IEEE/CVF international conference on
  computer vision}, 1588--1597.

\bibitem[{Wang et~al.(2019)Wang, Sun, Liu, Sarma, Bronstein, and
  Solomon}]{wang2019dynamic}
Wang, Y.; Sun, Y.; Liu, Z.; Sarma, S.~E.; Bronstein, M.~M.; and Solomon, J.~M.
  2019.
\newblock Dynamic graph cnn for learning on point clouds.
\newblock \emph{ACM Transactions on Graphics (tog)}, 38(5): 1--12.

\bibitem[{Wu et~al.(2025)Wu, Wan, Fu, Pfrommer, Zhong, Zheng, Zhang, and
  Beyerer}]{wu2025samble}
Wu, C.; Wan, Y.; Fu, H.; Pfrommer, J.; Zhong, Z.; Zheng, J.; Zhang, J.; and
  Beyerer, J. 2025.
\newblock SAMBLE: Shape-Specific Point Cloud Sampling for an Optimal Trade-Off
  Between Local Detail and Global Uniformity.
\newblock In \emph{Proceedings of the Computer Vision and Pattern Recognition
  Conference}, 1342--1352.

\bibitem[{Wu et~al.(2023)Wu, Zheng, Pfrommer, and Beyerer}]{wu2023attention}
Wu, C.; Zheng, J.; Pfrommer, J.; and Beyerer, J. 2023.
\newblock Attention-based point cloud edge sampling.
\newblock In \emph{Proceedings of the IEEE/CVF Conference on Computer Vision
  and Pattern Recognition}, 5333--5343.

\bibitem[{Wu et~al.(2015)Wu, Song, Khosla, Yu, Zhang, Tang, and
  Xiao}]{wu20153d}
Wu, Z.; Song, S.; Khosla, A.; Yu, F.; Zhang, L.; Tang, X.; and Xiao, J. 2015.
\newblock 3d shapenets: A deep representation for volumetric shapes.
\newblock In \emph{Proceedings of the IEEE conference on computer vision and
  pattern recognition}, 1912--1920.

\bibitem[{Xu et~al.(2021)Xu, Tang, Zhu, Sun, and Pu}]{xu2021sgmnet}
Xu, J.; Tang, X.; Zhu, Y.; Sun, J.; and Pu, S. 2021.
\newblock Sgmnet: Learning rotation-invariant point cloud representations via
  sorted gram matrix.
\newblock In \emph{Proceedings of the IEEE/CVF International Conference on
  Computer Vision}, 10468--10477.

\bibitem[{Yi et~al.(2016)Yi, Kim, Ceylan, Shen, Yan, Su, Lu, Huang, Sheffer,
  and Guibas}]{yi2016scalable}
Yi, L.; Kim, V.~G.; Ceylan, D.; Shen, I.-C.; Yan, M.; Su, H.; Lu, C.; Huang,
  Q.; Sheffer, A.; and Guibas, L. 2016.
\newblock A scalable active framework for region annotation in 3d shape
  collections.
\newblock \emph{ACM Transactions on Graphics (ToG)}, 35(6): 1--12.

\bibitem[{Yu et~al.(2024)Yu, Hou, Qin, Saleh, Shugurov, Wang, Busam, and
  Ilic}]{yu2024riga}
Yu, H.; Hou, J.; Qin, Z.; Saleh, M.; Shugurov, I.; Wang, K.; Busam, B.; and
  Ilic, S. 2024.
\newblock Riga: Rotation-invariant and globally-aware descriptors for point
  cloud registration.
\newblock \emph{IEEE Transactions on Pattern Analysis and Machine
  Intelligence}, 46(5): 3796--3812.

\bibitem[{Yu et~al.(2023)Yu, Qin, Hou, Saleh, Li, Busam, and
  Ilic}]{yu2023rotation}
Yu, H.; Qin, Z.; Hou, J.; Saleh, M.; Li, D.; Busam, B.; and Ilic, S. 2023.
\newblock Rotation-invariant transformer for point cloud matching.
\newblock In \emph{Proceedings of the IEEE/CVF conference on computer vision
  and pattern recognition}, 5384--5393.

\bibitem[{Yu, Zhang, and Cai(2023)}]{yu2023rethinking}
Yu, J.; Zhang, C.; and Cai, W. 2023.
\newblock Rethinking rotation invariance with point cloud registration.
\newblock In \emph{Proceedings of the AAAI Conference on Artificial
  Intelligence}, volume~37, 3313--3321.

\bibitem[{Zhang et~al.(2023)Zhang, Yu, Zhang, and Cai}]{zhang2023parot}
Zhang, D.; Yu, J.; Zhang, C.; and Cai, W. 2023.
\newblock Parot: Patch-wise rotation-invariant network via feature
  disentanglement and pose restoration.
\newblock In \emph{Proceedings of the AAAI Conference on Artificial
  Intelligence}, volume~37, 3418--3426.

\bibitem[{Zhang et~al.(2019)Zhang, Hua, Rosen, and Yeung}]{zhang2019rotation}
Zhang, Z.; Hua, B.-S.; Rosen, D.~W.; and Yeung, S.-K. 2019.
\newblock Rotation invariant convolutions for 3d point clouds deep learning.
\newblock In \emph{2019 International conference on 3d vision (3DV)}, 204--213.
  IEEE.

\bibitem[{Zhang, Yang, and Xiang(2024)}]{zhang2024risurconv}
Zhang, Z.; Yang, L.; and Xiang, Z. 2024.
\newblock Risurconv: Rotation invariant surface attention-augmented
  convolutions for 3d point cloud classification and segmentation.
\newblock In \emph{European Conference on Computer Vision}, 93--109. Springer.

\bibitem[{Zhao et~al.(2022)Zhao, Yang, Xiong, Zhu, Cao, and
  Li}]{zhao2022rotation}
Zhao, C.; Yang, J.; Xiong, X.; Zhu, A.; Cao, Z.; and Li, X. 2022.
\newblock Rotation invariant point cloud analysis: Where local geometry meets
  global topology.
\newblock \emph{Pattern Recognition}, 127: 108626.

\bibitem[{Zhou et~al.(2021)Zhou, Feng, Fang, Wei, Qin, and
  Lu}]{zhou2021adaptive}
Zhou, H.; Feng, Y.; Fang, M.; Wei, M.; Qin, J.; and Lu, T. 2021.
\newblock Adaptive graph convolution for point cloud analysis.
\newblock In \emph{Proceedings of the IEEE/CVF international conference on
  computer vision}, 4965--4974.

\end{thebibliography}

\end{document}